\documentclass[11pt]{article}

\usepackage[final]{acl}

\usepackage{times}
\usepackage{latexsym}
\usepackage{graphicx}               % 插图
\usepackage{booktabs}               % 专业表格
\usepackage{caption}                % 图表标题
\usepackage{enumitem}

\usepackage{xcolor}
\definecolor{lightgray}{gray}{0.95}
\usepackage{listings}
\definecolor{jsonkey}{RGB}{0,0,180}
\definecolor{jsonstring}{RGB}{163,21,21}
\definecolor{jsoncomment}{RGB}{0,128,0}

\lstdefinelanguage{json}{
  basicstyle=\ttfamily\footnotesize,
  showstringspaces=false,
  numbers=left,
  numberstyle=\tiny,
  stepnumber=1,
  numbersep=5pt,
  breaklines=true,
  backgroundcolor=\color{lightgray},
  literate=
    *{0}{{{\color{jsonstring}0}}}{1}
     {1}{{{\color{jsonstring}1}}}{1}
     {2}{{{\color{jsonstring}2}}}{1}
     {3}{{{\color{jsonstring}3}}}{1}
     {4}{{{\color{jsonstring}4}}}{1}
     {5}{{{\color{jsonstring}5}}}{1}
     {6}{{{\color{jsonstring}6}}}{1}
     {7}{{{\color{jsonstring}7}}}{1}
     {8}{{{\color{jsonstring}8}}}{1}
     {9}{{{\color{jsonstring}9}}}{1}
     {:}{{{\color{jsonkey}:}}}{1}
     {,}{{{\color{jsonkey},}}}{1}
     {\{}{{{\color{jsonkey}\{}}}{1}
     {\}}{{{\color{jsonkey}\}}}}{1}
     {[}{{{\color{jsonkey}[}}}{1}
     {]}{{{\color{jsonkey}]}}}{1}
}

\lstdefinestyle{prompt}{
  basicstyle=\ttfamily\small,
  showstringspaces=false,
  breaklines=true,
  frame=tb,
  framerule=0.5pt,
  rulecolor=\color{black!30},
  backgroundcolor=\color{white}
}
\usepackage[T1]{fontenc}
\usepackage[utf8]{inputenc}

\usepackage{microtype}

\usepackage{inconsolata}

\usepackage{graphicx}
\usepackage{algorithm}
\usepackage{algorithmic}
\usepackage{amsmath}
\usepackage{subcaption}
\usepackage[table]{xcolor}
\usepackage{multirow}
\definecolor{lightgray}{gray}{0.95}
\usepackage{listings}
\definecolor{jsonkey}{RGB}{0,0,180}
\definecolor{jsonstring}{RGB}{163,21,21}
\definecolor{jsoncomment}{RGB}{0,128,0}
\lstdefinelanguage{json}{
  basicstyle=\ttfamily\small,
  showstringspaces=false,
  numbers=left,
  numberstyle=\tiny,
  stepnumber=1,
  numbersep=5pt,
  breaklines=true,
  backgroundcolor=\color{lightgray},
  literate=
    *{0}{{{\color{jsonstring}0}}}{1}
     {1}{{{\color{jsonstring}1}}}{1}
     {2}{{{\color{jsonstring}2}}}{1}
     {3}{{{\color{jsonstring}3}}}{1}
     {4}{{{\color{jsonstring}4}}}{1}
     {5}{{{\color{jsonstring}5}}}{1}
     {6}{{{\color{jsonstring}6}}}{1}
     {7}{{{\color{jsonstring}7}}}{1}
     {8}{{{\color{jsonstring}8}}}{1}
     {9}{{{\color{jsonstring}9}}}{1}
     {:}{{{\color{jsonkey}:}}}{1}
     {,}{{{\color{jsonkey},}}}{1}
     {\{}{{{\color{jsonkey}\{}}}{1}
     {\}}{{{\color{jsonkey}\}}}}{1}
     {[}{{{\color{jsonkey}[}}}{1}
     {]}{{{\color{jsonkey}]}}}{1}
}
\title{Dual-Axis Generative Reward Model Toward Semantic and Turn-taking Robustness in Interactive Spoken Dialogue Models}

\author{
  \textbf{Yifu Chen\textsuperscript{1}\thanks{These authors contributed equally.}\quad
  Shengpeng Ji\textsuperscript{1}\footnotemark[1]\quad
  Zhengqing Liu\textsuperscript{1}\footnotemark[1]\quad
  Qian Chen\textsuperscript{2}\quad} \\
  \textbf{Wen Wang\textsuperscript{2}\quad
  Ziqing Wang\textsuperscript{3}\quad
  Yangzhuo Li\textsuperscript{1}\quad
  Tianle Liang\textsuperscript{1}\quad
  Zhou Zhao\textsuperscript{1}\thanks{Corresponding author.}} \\
  \\
  $^{\text{1}}$~Zhejiang University \quad
  $^{\text{2}}$~Tongyi Fun Team, Alibaba Group \quad
  $^{\text{3}}$~Beijing University of Technology \\
  \texttt{22551267@zju.edu.com, zhaozhou@zju.edu.cn}
}

\begin{document}
\maketitle
\begin{abstract}
Achieving seamless, human-like interaction remains a key challenge for full-duplex spoken dialogue models (SDMs). Reinforcement learning (RL) has substantially enhanced text- and vision-language models, while well-designed reward signals are crucial for the performance of RL. We consider RL a promising strategy to address the key challenge for SDMs. However, a fundamental barrier persists: prevailing automated metrics for assessing interaction quality rely on superficial proxies, such as behavioral statistics or timing-prediction accuracy, failing to provide reliable reward signals for RL. On the other hand, human evaluations, despite their richness, remain costly, inconsistent, and difficult to scale. We tackle this critical barrier by proposing a \textbf{Dual-Axis Generative Reward Model}, which is trained to understand complex interaction dynamics using a detailed taxonomy and an annotated dataset, produces a single score and, crucially, provides separate evaluations for \textbf{semantic quality} and \textbf{interaction timing}. Such dual outputs furnish precise diagnostic feedback for SDMs and deliver a dependable, instructive reward signal suitable for online reinforcement learning. Our model achieves state-of-the-art performance on interaction-quality assessment across a wide spectrum of datasets, spanning synthetic dialogues and complex real-world interactions.Our page could be found at \url{https://github.com/MM-Speech/DualAxisRM}.

\end{abstract}

\section{Introduction}
% The pursuit of creating Spoken Dialogue Models~\cite{xu2025qwen2,ding2025kimi,chen2025minmo,fang2025llama} that can engage in natural, fluid, and contextually aware conversations is a central goal in artificial intelligence~\cite{ji2024wavchat}. While significant strides have been made in language understanding and generation, achieving human-like interactional fluency, marked by seamless turn-taking and free-form responsive behaviors, remains a critical and unresolved challenge~\cite{lin2025full,zhang2024omniflatten}. Effective interaction management is a multi-stage cognitive process that involves perceiving user intent, deciding when to speak or listen, and responding coherently. Failures in any of these stages can significantly degrade the user experience, leading to perceptions of artificiality and frustration. Therefore, capturing the dual requirements of what to say (semantic quality) and when to say (timing correctness) is essential for developing the next generation of interactive spoken dialogue models.

%%%%The pursuit of developing Spoken Dialogue Models (SDMs)~\cite{xu2025qwen2,ding2025kimi,chen2025minmo,fang2025llama} that can engage in natural, human-like, and contextually aware conversations is a central goal in artificial intelligence~\cite{ji2024wavchat}. 
There have been significant advancements in Spoken Dialogue Models (SDMs) toward natural, human-like, and contextually aware conversations~\cite{xu2025qwen2,ding2025kimi,chen2025minmo,fang2025llama,ji2024wavchatsurveyspokendialogue,chen-etal-2025-wavrag,li2026wavbench}.
A key challenge for SDMs is achieving smooth interaction, which requires mastering not only what to say (semantic quality) but also when to say it (timing correctness), especially in free-form full-duplex conversations~\cite{lin2025full, zhang2024omniflatten}.
 Reinforcement Learning (RL)~\cite{cao2024survey} offers a powerful framework for optimizing such interactive behaviors, but its application in SDMs is hindered by the absence of a \textit{reliable}, \textit{low-latency} reward signal. Human evaluation, the typical gold standard, has been effectively used in offline methods like Direct Preference Optimization (DPO) to enhance full-duplex models~\cite{rafailov2024directpreferenceoptimizationlanguage, yu2025salmonn, wu2025aligning, veluri2024beyond}; however, its practical application in RL is limited by its high cost, inconsistency, and poor scalability.
 %%%%, rendering it unsuitable for online reinforcement learning approaches~\cite{retzlaff2024human} which usually has a higher optimization performance ceiling.
\begin{figure*}[t]
\centering
\includegraphics[width=0.8\linewidth]{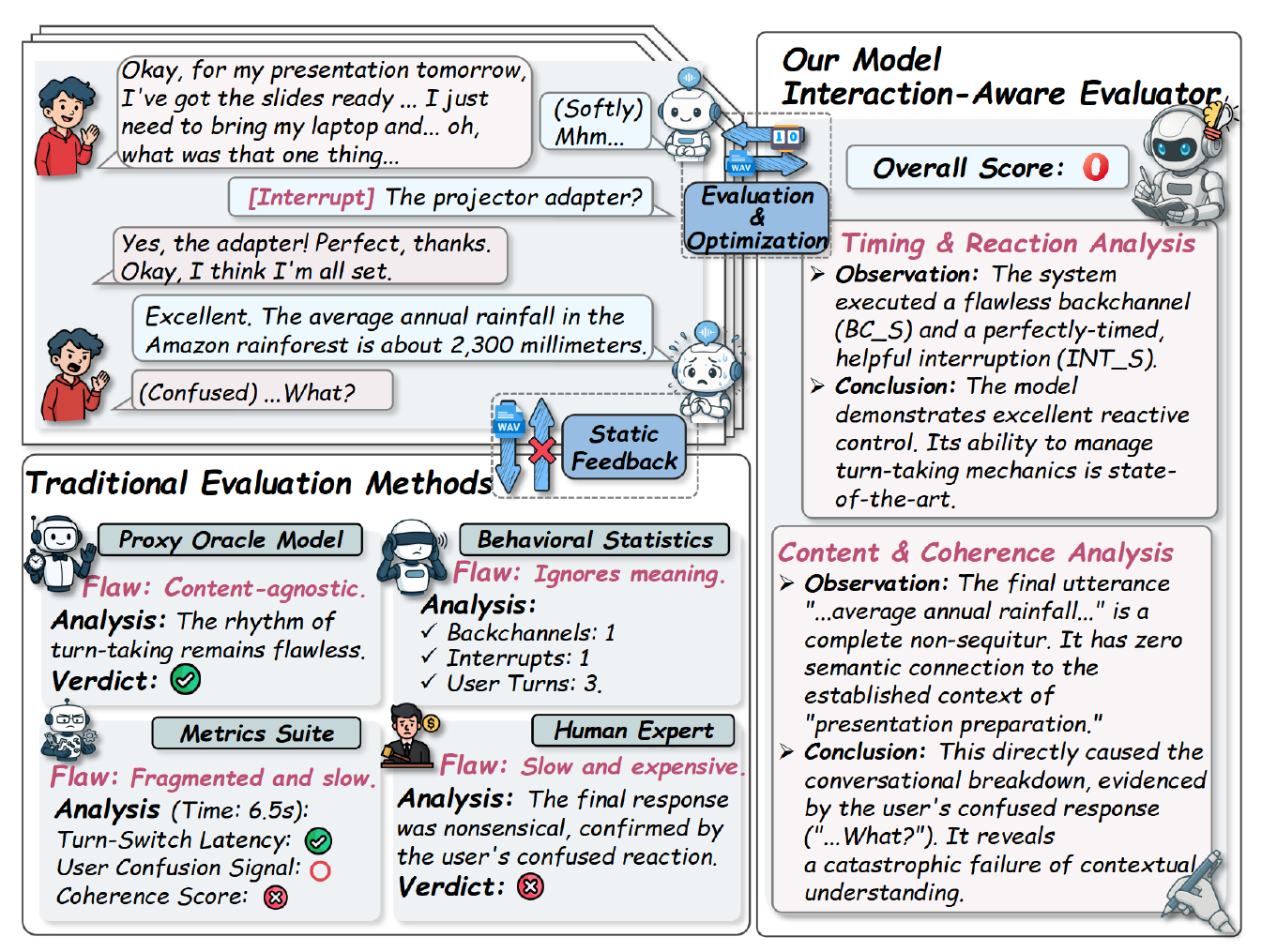} 
\caption{Comparison between our interaction-aware evaluator and traditional evaluation methods for interaction quality.}
\label{fig1}
\end{figure*}

To reduce reliance on costly human label, various automated proxies have been proposed. As summarized in Figure~\ref{fig1}, current automatic evaluation methods often capture only one side of interaction quality. \textit{Oracle-aligned proxies}~\cite{arora2025talking} emphasize frame-level interaction timing and—being relatively content-insensitive—may rate an interaction as well-timed even when semantic breakdowns occur. \textit{Behavioral statistics}~\cite{defossez2024moshi,wang2025ntpp} count surface events but do not assess their appropriateness or coherence. \textit{Metrics suites}~\cite{lin2025full} provide fine-grained diagnostics, yet they yield fragmented, high-latency signals that are difficult to fuse into a unified, real-time reward; therefore, these proxies are often ill-suited to the high-frequency feedback loop required for online RL.
A promising recent direction is the ``LLM-as-a-judge'' paradigm~\cite{gu2024survey, chiang2025audioawarelargelanguagemodels, ji2025wavreward}. Yet, this approach faces a critical challenge in full-duplex interactions, where simultaneous speaking and listening create complex dynamics. Existing Large Language Model (LLM) evaluators, designed for simpler turn-based dialogues, lack the perception grounding to assess the crucial interplay of when to speak and what to say.  As our case study in Figure~\ref{case} demonstrates, even state-of-the-art (SOTA) commercial models would mischaracterize critical turn-taking violations, incorrectly labeling a flawed interaction as perfectly ``smooth''. This limitation motivates the core objective of our work—to \textbf{develop an automated LLM-based reward model that well approximates expert human judgment for complex full-duplex interactions}, which performs with \textbf{high consistency and interpretability as well as low latency}, essential for RL optimization of SDMs.

We introduce an \textbf{interaction-aware LLM-based reward model} grounded in a systematic taxonomy of interaction dynamics and the common failure modes and trained on a hybrid suite of synthetic and real-world full-duplex interaction datasets, with a broad coverage of the failure taxonomy and interaction dynamics.
%%%%First, we synthesize a \textbf{146-hour dataset} engineered for broad coverage of the failure taxonomy to support initial supervised training. This is supplemented by a \textbf{12-hour real-world human-human (RW-HH) set} from Seamless Interaction~\cite{agrawal2025seamlessinteractiondyadicaudiovisual} for gold-standard testing, and \textbf{10-hour real-world human-machine (RW-HM) sets}. One 8-hour RW-HM set is used for fine-tuning on authentic machine-induced errors, while the other is reserved for final evaluation. All datasets feature precise \wen{human?} annotations of interaction quality, failure types, transcriptions, and timestamps.  
We train a foundation LLM through a progressive three-stage training paradigm: (i) supervised grounding in dual-track audio structure to familiarize the model with the fundamental characteristics and events of dual-track dialogues, (ii) distillation of Chain-of-Thought (CoT) reasoning to bootstrap its analytical abilities on assessing interactions, and (iii) enhancing generalization and capabilities via Group Relative Policy Optimisation~\cite{shao2024deepseekmathpushinglimitsmathematical}.  The resulting reward model generates interpretable, multi-faceted assessments with a single reward score. Its structured output enables a key diagnostic capability: decoupling the assessment of the SDM's \textit{reactive control (the mechanics of turn management and timing)} from its \textit{content generation (semantic relevance and coherence)} and delivering a reliable, instructive reward signal suitable for RL optimization. Our main contributions are as follows:
\begin{itemize}[leftmargin=*,noitemsep]
    \item \textbf{A Novel Dual-Axis Reward Model}: We introduce an interaction-aware generative reward model that provides structured, interpretable feedback by explicitly decoupling the assessment of semantic coherence from interaction timing. 
    \item \textbf{A Training Recipe}: We establish a formal taxonomy of dyadic interaction dynamics and common interaction failure modes, investigate weaknesses of current SOTA LLMs for interaction-quality evaluation, and based on this, design a recipe that trains audio comprehension models into high-performing interaction-quality reward models.
    \item \textbf{New Annotated Datasets}: We construct and annotate novel datasets, comprising a large corpus of synthetic dialogues, authentic human-machine and human-human interactions.
\end{itemize}
\begin{figure*}[t]
\centering
\includegraphics[width=0.8\textwidth]{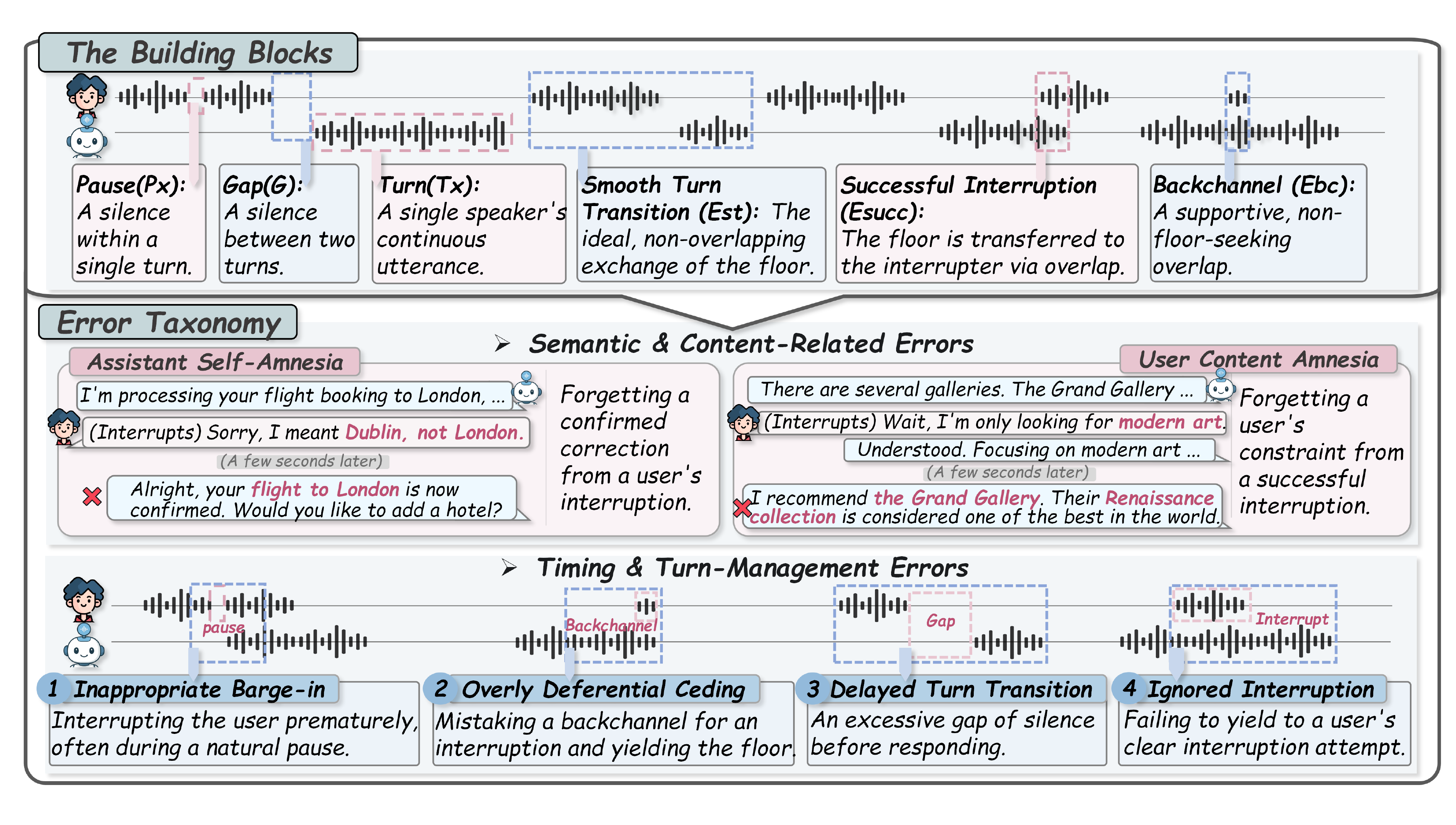} % Reduce the figure size so that it is slightly narrower than the column.
\caption{Formal taxonomy of interaction dynamics and failure types.}
\label{fig2}
\end{figure*}

\section{Related Work}
\noindent\textbf{Reward Modeling for Full-Duplex Interaction.} The advent of ``LLM-as-a-Judge'' has advanced automated evaluation in the speech domain, with models assessing qualities from speaking styles~\cite{chiang2025audioawarelargelanguagemodels,lu2026modelingbenchmarkingspokendialogue} to conversational effectiveness~\cite{yazdani2025evaluating, chen2025audiolargelanguagemodels}. These judgments increasingly serve as supervision for generative reward models like WavReward~\cite{ji2025wavreward,ji2024wavtokenizer}, which provide structured feedback for spoken dialogue systems~\cite{lin2024align}. However, these evaluators are almost exclusively designed for \textbf{mono-channel, turn-based dialogues}, making them unable to assess the critical temporal dynamics of full-duplex interaction.
When applied to optimizing full-duplex SDMs, this limitation creates a dichotomy in current RL approaches. Online RL methods like ORISE~\cite{chen2025reinforcement} rely on rule-based heuristics (e.g., from VAD), which capture timing cues but provide coarse, semantically-unaware feedback. Conversely, offline, preference-based methods like DPO~\cite{wu2025aligning} focus on aligning semantic content but utilize latent, high-cost feedback signals that ignore fine-grained interaction timing. Consequently, the field lacks a reward mechanism capable of \textbf{jointly and synchronously evaluating what is said (semantic appropriateness) and when it is said (temporal correctness)} in a manner suitable for online policy optimization. Our work addresses this critical gap by introducing the first decoupled generative reward model designed specifically for RL for free-form full-duplex interactions. Note that while structured error taxonomies for text-based chat-oriented dialogue have been proposed~\cite{finch-etal-2023-dont}, our taxonomy is grounded in full-duplex timing and turn-taking structures (overlaps, gaps, interruptions, backchannels) combined with content errors in spoken dialogues, making it complementary to prior text-focused work.

\section{Methodology}
\label{sec:methodology}

\subsection{Taxonomy of Interaction}
\label{ssec:modeling_and_errors}

To systematically analyze conversational interaction behaviors, we first establish a formal model of interaction dynamics as shown in Figure~\ref{fig2}. The elemental feature is the phonatory state for each interlocutor \(X_i\), captured by the function \(\sigma(X_i, t) \in \{\text{Speech}, \text{Silence}\}\). These states form contiguous maximal intervals: speech segments (\(\mathcal{S}_{X_i}\)) and silence segments (\(\mathcal{I}_{X_i}\)). From these, we define the structural units of interaction by critically distinguishing silences internal to a speaker's contribution from those between speakers. A \textbf{pause} (\(\mathcal{P}_{X_i}\)) is an \textit{intra-speaker} silence—a segment \(\mathcal{I}_{X_i}\) preceded and followed by speech from the \emph{same} speaker. A speaker's \textbf{turn} (\(\mathcal{T}_{X_i}\)) is therefore a maximal period of their speech, potentially comprising multiple speech segments separated by their own pauses. In contrast, a \textbf{gap} (\(\mathcal{G}\)) is an \textit{inter-speaker} silence that facilitates a speaker change, defined as a period of mutual silence between one speaker's turn ending and another speaker's beginning. Its counterpart, an \textbf{overlap} (\(\mathcal{O}\)), is a period of simultaneous speech.

The temporal arrangement of these structural units gives rise to key functional interaction events. The canonical, non-overlapping exchange of the conversational floor is the \textbf{Smooth Turn Transition} ($E_{\text{st}}$), defined by the sequence $\mathcal{T}_{X_i} \rightarrow \mathcal{G} \rightarrow \mathcal{T}_{X_j}$ where $i \neq j$. More complex events involving simultaneous speech, or overlaps ($\mathcal{O}$), are classified based on their resolution, which is determined by which speaker holds the floor after the overlap concludes. A clear transfer of the floor from the original speaker ($X_i$) to the interrupting speaker ($X_j$) defines a \textbf{Successful Interruption} ($E_{\mathrm{succ}}$). Structurally, this occurs when $X_i$ yields by falling silent, while $X_j$ continues their speech to capture the turn. Conversely, in scenarios where the original speaker ($X_i$) retains the floor, the classification is further determined by the function of the interrupting utterance ($\mathcal{S}_{X_j}$). If this utterance is a short, non-competitive signal of listenership or agreement (e.g., ``uh-huh'', ``right''), the event is classified as a \textbf{Backchannel} ($E_{\mathrm{bc}}$). However, if the utterance represents a more substantial and competitive content for the conversational floor, which the original speaker successfully defends by continuing to speak, the event is categorized as a \textbf{Failed Interruption} ($E_{\mathrm{fail}}$).

Then we define an interaction error as any systematic departure from
the timing and semantic norms observed in large human–
human and human–machine corpora. To empirically ground our taxonomy, we first analyze two complementary sources:
(i) publicly-available conversation datasets; and (ii) in-house annotations of 10h of
human–machine dialogues spanning three SOTA models. Our taxonomy organizes these empirically grounded
failures along two axes: errors in semantic coherence and in turn-management.

\textbf{Semantic and Content-Related Errors} encompass failures in the semantic and pragmatic integrity of the conversation. The primary manifestation is \textbf{Contextual Incoherence}, where an utterance lacks logical consistency or relevance to the preceding context. A particularly critical example of this is \textbf{Interruption Amnesia}, which refers to the system’s failure to preserve and update the discourse state after an interruption.

\textbf{Timing and Turn-Management Errors} reflect a mechanical failure in the \emph{when} of an utterance. These can be subdivided by the system’s reaction speed. Over-reactive errors include \emph{Inappropriate Barge-in}, where the system initiates an overlap (\(\mathcal{O}\)) that is neither supportive nor a justified interruption, often by misinterpreting a user's natural pause (\(\mathcal{P}\)) as a turn-yielding cue. Another is \emph{Overly Deferential Ceding}, where the system incorrectly treats a user's supportive backchannel (\(E_{\mathrm{bc}}\)) as a full interruption attempt and prematurely terminates its own turn, leading to a hesitant flow. Conversely, under-reactive errors demonstrate a failure to respond promptly. \emph{Delayed Turn Transition} manifests as an excessively long gap (\(\mathcal{G}\)) in what should be a smooth turn transition (\(E_{\text{st}}\)), creating awkward silence. Its counterpart, \emph{Ignored Interruption}, occurs when the system fails to yield its turn in response to a clear user attempt to take the floor, transforming what should have been a successful interruption (\(E_{\mathrm{succ}}\)) into a prolonged and frustrating failed interruption (\(E_{\mathrm{fail}}\)).
\begin{figure*}[t]
\centering
\includegraphics[width=\textwidth]{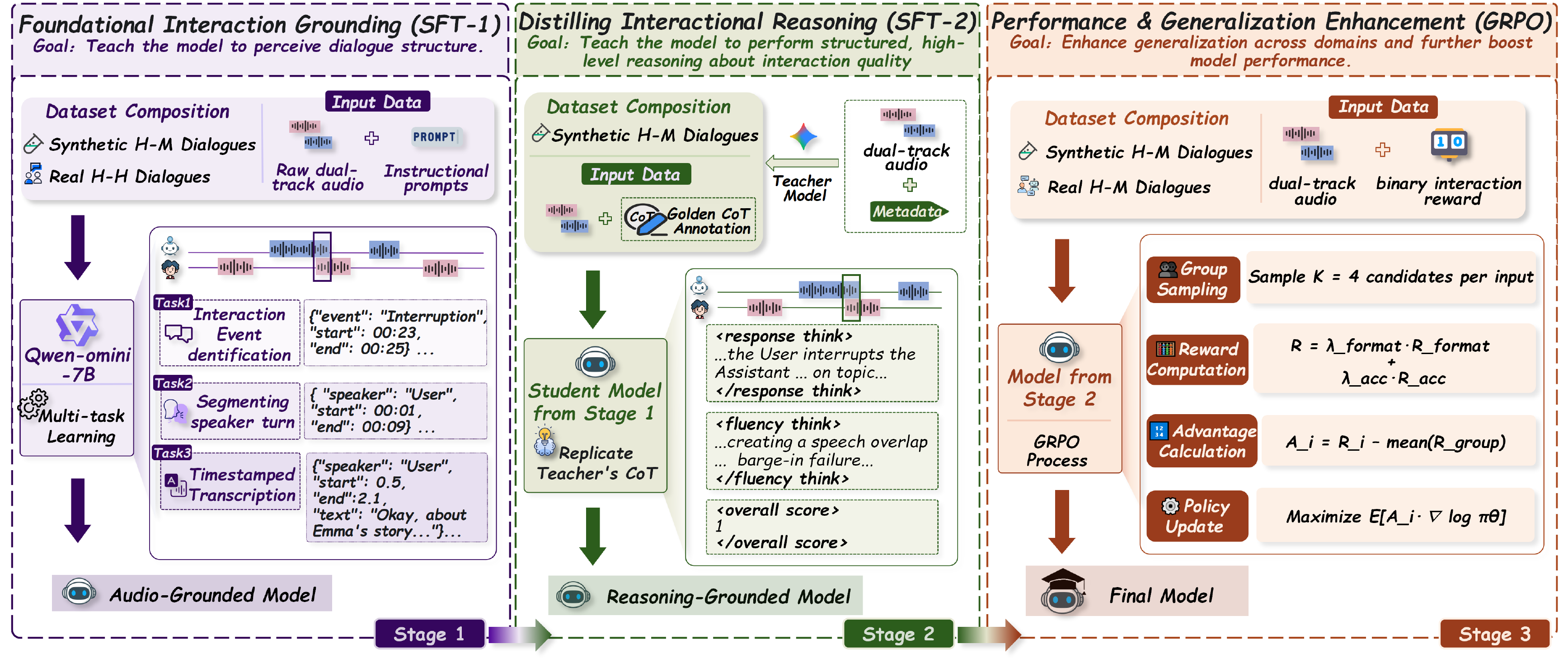} 
\caption{Overview of Training Stages.}
\label{fig:training-stages}
\end{figure*}
\subsection{Dual-Axis Reward Model}
To robustly evaluate an SDM on the aspects defined above, we frame our solution as the development of a decoupled reward model. In the context of reinforcement learning, a reward model, denoted as \(R_\theta\), a separate machine learning model that estimates a numerical reward score for an agent's output based on its alignment with desired behavior. We formalize our specific task as learning a reward model that maps a fully specified interaction scenario, \(x\), to a structured evaluation tuple. The model's function is therefore:
\[
R_\theta(x) = (\text{CoT}_{\text{sem}}, \text{CoT}_{\text{turn}}, S)
\]
Here, the model's output consists of three components: two parallel \textbf{Chain-of-Thought (CoT)} analyses and an overall \textbf{binary score}, \(S\), which serves as the final reward signal. The first chain-of-thought focuses on \textbf{Semantic Coherence} (\(CoT_{\text{sem}}\)), explicitly reasoning about the \emph{content}: Was the SDM’s response appropriate and logically consistent given the context and the user’s input? The other chain focuses on \textbf{Turn Management} (\(CoT_{\text{turn}}\)), examining the \emph{interaction timing}: Did the SDM follow the turn-taking norms, e.g., not interrupt improperly, respond without undue delay, correctly handle any barge in? By separating these two threads, the model can make transparent, criterion-specific judgments before arriving at an integrated conclusion. This design is inspired by the success of chain-of-thought reasoning in complex tasks – breaking down the evaluation into interpretable steps helps ensure no aspect is overlooked and mirrors how human annotators might separately consider \emph{what} was said versus \emph{how} it was said.

Following this dual analysis, the model renders the final, integrated \textbf{Binary Correctness Score} \(S \in \{0, 1\}\). A score of 1 is assigned if and only if the SDM's performance is deemed acceptable across both the semantic and turn-taking dimensions; a failure in either dimension results in a score of 0. We opted for this dichotomous rating over a graded scale for three primary reasons. First, it aligns with the nature of real-world user feedback and modern training paradigms; for instance, Reinforcement Learning from Human Feedback (RLHF) fundamentally distills preferences into binary win/loss signals for training \cite{han2025reinforcement}. Second, binary annotation has been demonstrated to yield higher inter-annotator reliability and consistency than multi-point Likert scales, reducing subjective drift \cite{smith2022human}. Third, in user-facing full-duplex spoken dialogue systems, interactions that fail badly on either axis are typically judged as simply ``bad'' by users regardless of which axis failed; the binary objective matches this holistic success criterion and provides a clear, actionable objective for the model: to maximize the probability of achieving a holistically successful interaction where both content and timing are correct \cite{guo2024multi}.

\subsection{Three-Stage Training Recipe}
\label{ssec:training_recipe}

Our methodology employs a hybrid dataset of synthetic and real-world data. 
The synthetic portion is engineered for broad coverage of our failure taxonomy through a controlled pipeline. 
First, we rewrite text dialogues to script specific interactional dynamics, explicitly controlling the timing of conversational turns. 
Subsequently, advanced TTS models (GPT-4o-mini-tts\footnote{https://platform.openai.com/docs/models/gpt-4o-mini-tts} and Gemini TTS\footnote{https://ai.google.dev/gemini-api/docs/speech-generation}) are utilized for fine-grained audio synthesis. 
This enables precise, programmatic control over temporal features such as \textbf{overlap durations} and \textbf{silence lengths}, with all events being \textbf{programmatically timestamped}.
The complete train dataset comprises 6,361 synthetic samples (approx. 146 hours), supplemented by real-world dialogues, including 100 human-human samples from Seamless Interaction~\cite{agrawal2025seamlessinteractiondyadicaudiovisual} and 289 of our own collected human-machine samples (approx. 10 hours). 
While the synthetic data is programmatically annotated, the collected real-world samples underwent \textbf{rigorous manual annotation by human experts}.
This process ensures all samples are richly labeled with speaker timestamps, transcripts, interaction quality, interaction events, and error types. 
Details of the data construction and annotation protocols are in Appendix~\ref{sec:dataset}.
We design a progressive three-stage training paradigm, as shown in Figure~\ref{fig:training-stages}.
\begin{table*}[!ht]
\centering
\small
% Helper command for creating two-line column headers (Dataset name + Acc. / F1)
\newcommand{\myheader}[1]{\begin{tabular}{@{}c@{}}\textbf{#1} \\ \small Acc. / F1\end{tabular}}

% Using tabular* to fit the table to \textwidth and fill extra space between columns
\begin{tabular*}{\textwidth}{@{\extracolsep{\fill}}lccccc}
\toprule
% \multirow correctly spans two rows and is now vertically centered.
\multirow{3}{*}[-0.6ex]{\textbf{ Model}} & \multicolumn{3}{c}{\textbf{Synthetic Data}} & \multicolumn{2}{c}{\textbf{Real-World Data}} \\
\cmidrule(lr){2-4} \cmidrule(lr){5-6}
& \myheader{In-Distribution} & \myheader{Fine-Grained} & \myheader{OOD} & \myheader{RW-HH} & \myheader{RW-HM} \\
\midrule
\rowcolor{gray!20}\multicolumn{6}{l}{\itshape Primary Comparison} \\
\textbf{Ours} & \textbf{0.9853} / \textbf{0.9852} & \textbf{0.8500} / \textbf{0.8476} & \textbf{0.9614} / \textbf{0.9610} & \textbf{\underline{0.8679}} / \textbf{0.6931} & \textbf{0.7727} / \textbf{0.7647} \\
\textit{QwenOmni (base)} & 0.0733 / 0.0811 & 0.2500 / 0.2500 & 0.0858 / 0.0837 & 0.4906 / 0.4851 & 0.5227 / 0.5227 \\
\midrule
\rowcolor{gray!20}\multicolumn{6}{l}{\itshape Open-Source Models} \\
Qwen2Audio & 0.2454 / 0.2431 & 0.3000 / 0.2917 & 0.0944 / 0.0911 & 0.4528 / 0.4501 & 0.3636 / 0.3604 \\
AudioReasoner & 0.3004 / 0.2831 & 0.2500 / 0.2500 & 0.3262 / 0.3105 & 0.7170 / 0.4186 & 0.5909 / 0.5909 \\
KimiAudio & 0.4432 / 0.4248 & 0.3500 / 0.3417 & 0.5408 / 0.5310 & 0.7736 / 0.4915 & 0.4318 / 0.4286 \\
Audio-Flamingo3 & 0.2967 / 0.2825 & 0.2500 / 0.2500 & 0.3176 / 0.2995 & 0.6604 / 0.4241 & 0.4318 / 0.4286 \\
\midrule
\rowcolor{gray!20}\multicolumn{6}{l}{\itshape Closed-Source Models} \\
GPT-4o & 0.5604 / 0.5517 & 0.3750 / 0.2699 & 0.5708 / 0.5539 & \textbf{0.9245} / 0.6467 & 0.6818 / 0.6818 \\
Gemini-2.5-Pro & 0.7473 / 0.7423 & 0.6750 / 0.6791 & 0.6781 / 0.6715 & 0.6604 / 0.4241 & 0.6136 / 0.6136 \\
Gemini-2.5-Flash & 0.6740 / 0.6685 & 0.4500 / 0.4472 & 0.6481 / 0.6402 & 0.7736 / 0.4915 & 0.6818 / 0.6818 \\
\bottomrule
\end{tabular*}
\caption{\textbf{Accuracy and Macro F1-Score comparison} between our model (\textbf{Ours}, Qwen-2.5-Omni-7B pos-trained with the three stages) and all open-source and closed-source baselines (baselines all in zero-shot setting) across all evaluation datasets.}
\label{tab:comprehensive_results}
\end{table*}

\noindent \textbf{Stage 1: Foundational Interaction Grounding.}
This initial stage addresses a key limitation of standard pre-trained models, Qwen-2.5-Omni-7B~\cite{xu2025qwen2}: its inability to comprehend the structural and specific interaction events of dual-track spoken dialogues. These models, typically pre-trained on single-track audio or text, lack explicit representations for conversational phenomena like silence or overlapping speech. To address this, we employ Supervised Fine-Tuning (SFT) with multi-tasks, pairing each audio sample with a specific instruction to guide the model on a single, designated task. Conditioned on distinct prompts, the model is trained to perform various analyses: identifying interaction events based on our classification and outputting their precise timestamps; performing speaker diarization by detecting and timestamping all speech segments attributed to specific speakers; or generating a complete diarized transcript with speaker-attributed utterances and timestamps.

\noindent \textbf{Stage 2: Distilling Interactional Reasoning.}
Building on the perceptual grounding from Stage 1, this stage trains the model to perform structured, high-level reasoning about interaction quality. We sample various interaction types from our synthetic dialogue dataset and \textbf{input their text metadata},including timestamps, transcription and speaker, into Gemini-2.5-pro to generate verifiably reliable structured reasoning and scores. The model checkpoint from Stage 1 is then fine-tuned on a 2,670-sample Chain-of-Thought (CoT) dataset. This process trains the model to generate a structured evaluation that decouples its assessment into two components: (1) \textit{Response Relevance}, analyzing the semantic coherence of an utterance, and (2) \textit{Interactional Fluency}, assessing interaction timing appropriateness.

\noindent \textbf{Stage 3: RL for Performance and Generalization Enhancement.}
To enhance the model's robustness against unseen interaction errors and move beyond supervised performance, we fine-tune the Stage-2 checkpoint using Group Relative Policy Optimisation(GRPO).

In this stage, for each dialogue \(x\), the current policy \(\pi_{\theta}\) samples a group of \(K\) candidate evaluations \(\{O_i\}_{i=1}^{K}\). Each candidate is assessed with a scalar reward \(r(x, O_i)\), which is a weighted combination of format adherence and score accuracy:
\[
r(x,O_i)=\lambda_{\mathrm{fmt}}\,I_{\mathrm{fmt}}(O_i)+\lambda_{\mathrm{acc}}\,I_{\mathrm{acc}}(O_i).
\]
Here, \(I_{\mathrm{fmt}}\) is a binary indicator that checks for the presence of the mandatory \texttt{<response think>}, \texttt{<fluency think>}, and \texttt{<score>} blocks. Similarly, \(I_{\mathrm{acc}}\) is a binary indicator that equals 1 if the extracted score matches the ground-truth label \(s_{\text{gt}} \in \{0,1\}\). The weights are constrained by \(\lambda_{\mathrm{fmt}}+\lambda_{\mathrm{acc}}=1\).

To stabilize training, GRPO normalize these rewards across the group of candidates. This yields the group-relative advantage \(\hat{A}_i\):
\[
\hat{A}_i = \frac{r(x, O_i) - \mu_r}{\sigma_r},
\]
where \(\mu_r\) and \(\sigma_r\) are the mean and standard deviation of the rewards \(\{r(x,O_j)\}_{j=1}^{K}\) in the group.

The GRPO objective maximizes a clipped version of the advantage. We define the importance sampling ratio as \(w_i(\theta) = \frac{\pi_{\theta}(O_i \mid x)}{\pi_{\theta_{\mathrm{old}}}(O_i \mid x)}\), where \(\pi_{\theta_{\mathrm{old}}}\) is the policy before the update. The objective function to be maximized is:
\begin{equation}
\resizebox{0.87\columnwidth}{!}{%
  $\displaystyle % To maintain display style math
  \begin{aligned}
  J_{\mathrm{GRPO}}(\theta)
  &= \underset{x, \{O_i\} \sim \pi_{\theta_{\text{old}}}}{\mathrm{E}}
    \Biggl[
      \frac{1}{K} \sum_{i=1}^{K}
      \min\Bigl( w_i(\theta)\hat{A}_i,\,
  \\[-0.3em]
  &\qquad\qquad
        \operatorname{clip}\bigl(w_i(\theta),\, 1-\varepsilon,\, 1+\varepsilon\bigr)\hat{A}_i
      \Bigr)
    \Biggr]
  \end{aligned}
  $
}
\end{equation}
 All training parameters and settings are detailed in the Appendix~\ref{train}.

\section{Experiments}

\subsection{Experimental Setup}

\noindent \textbf{Baselines.} Our model was benchmarked against a variety of strong baseline models. For closed-source models, we compared against \textbf{GPT-4o}~\footnote{https://platform.openai.com/docs/models/gpt-4o-audio-preview},  \textbf{Gemini-2.5-Pro} and \textbf{Gemini-2.5-Flash}~\footnote{https://ai.google.dev/gemini-api}. For open-source models, we included sota audio reason model: \textbf{Qwen2Audio}~\cite{chu2024qwen2}, \textbf{AudioReasoner}~\cite{xie2025audio}, \textbf{KimiAudio}~\cite{goel2025audio}, and \textbf{Audio-Flamingo3}. \textbf{Qwen2.5-Omni}~\cite{xu2025qwen2} served as our base model to establish a performance baseline.

\noindent \textbf{Evaluation Setup.} All models are evaluated in a zero-shot inference setting. The task is to predict the binary interaction quality score for each dialogue dual-track audio instance, guided by prompt detailed in Appendix~\ref{sec:inference_prompt}. The evaluation metric is \textbf{Accuracy and macro-F1 score}.

\noindent \textbf{Evaluation Datasets.}
 Our evaluation framework incorporates five distinct test sets, each designed to probe a different aspect of model performance. Detailed data statistics, construction details, and samples are provided in the Appendix~\ref{sec:realworld}.
\textbf{In-Distribution (ID):} This dataset is drawn from the same distribution as the training data. It serves as a fundamental benchmark to evaluate the model's core learning and inference capabilities on familiar data patterns.
 \textbf{Fine-Grained (FG) Analysis:} This dataset consists of 40 response pairs sampled from the ID test set. Each pair was manually analyzed and assigned a ground-truth label of either Correct Interaction or one of three specific failure types: Semantic Error, Over-reactive Errors, or Under-reactive Errors. This process allows for a precise calculation of the model's accuracy in identifying these targeted interaction phenomena
 \textbf{Out-of-Distribution (OOD):} This dataset is constructed from data sources distinct from those used in the training set, although the construction methodology remains analogous. It is designed to rigorously test the model's generalization ability when confronted with novel yet structurally similar conversational contexts.
\textbf{Real-World Human-Human (RW-HH):} This dataset is composed of annotated samples from the Seamless Interaction dataset. It provides a crucial test of the model's performance on authentic, unscripted human-human dialogues, evaluating its ability to comprehend the complex dynamics of naturalistic conversations.
 \textbf{Real-World Human-Machine (RW-HM):}  We recruited human annotators to collect real-world human–machine interactions data. It contains audio recordings of fluent English speakers interacting with state-of-the-art models, namely GPT-4o~\footnote{https://openai.com/index/hello-gpt-4o}, Gemini~\footnote{https://gemini.google/overview/gemini-live}, and Doubao~\footnote{https://www.doubao.com}. 
 
\begin{figure}[h!]
    \centering % 整体居中

    % 第一个子图
    \includegraphics[width=\columnwidth]{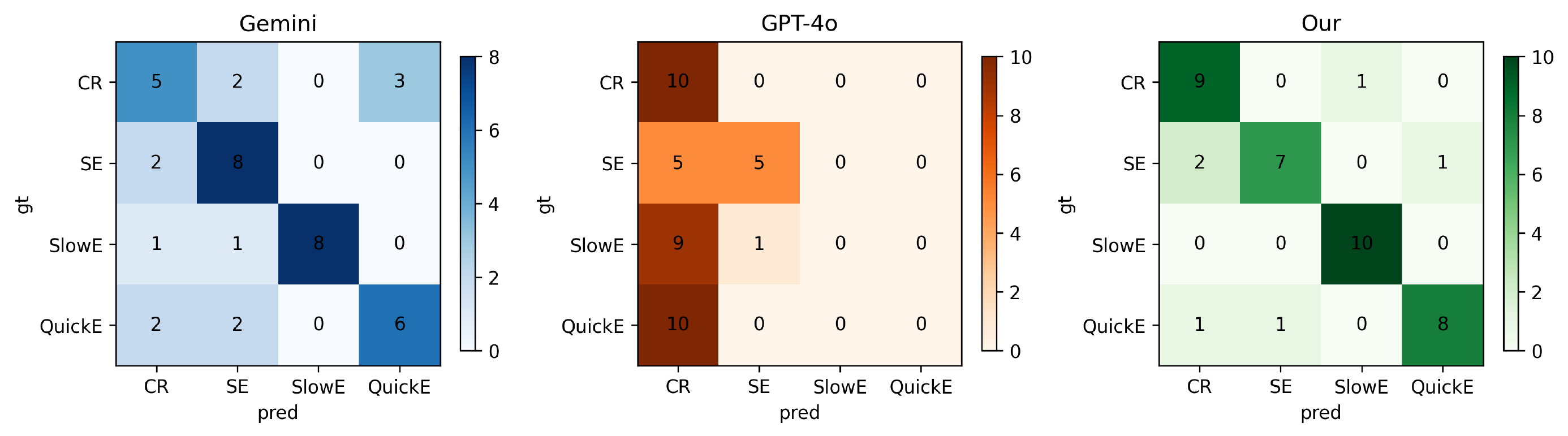}

    % 整个图的总标题和总标签
    \caption{Fine-grained error classification confusion matrix of different evaluators across interaction failure types: (a) Gemini-2.5-Pro, (b) GPT-4o, and (c) our model. Abbreviations: SE = Semantic Errors; QuickE = Over-reactive Errors (Inappropriate Barge-in, Overly Deferential Ceding); SlowE = Under-reactive Errors (Delayed Turn Transition, Ignored Interruption).}
    \label{fig:three_figures}
\end{figure}

\begin{table}[h!]
\centering
\resizebox{\columnwidth}{!}{%
\begin{tabular}{lcc}
\hline
\textbf{Model} & \textbf{Consistency Score} & \textbf{Win Rate } \\
\hline
\textbf{Ours} & \textbf{4.125} & \textbf{55\%} \\
Gemini-2.5-Pro             & 3.950 & 40\% \\
GPT-4o                     & 2.500 & 5\% \\
\hline
\end{tabular}
}
\caption{Evaluation of Chain-of-Thought (CoT) quality against human expert judgment. Our model leads substantially in both average consistency and preference win rate.}
\label{tab:cot_consistency}
\end{table}
\begin{table*}[h!]
\centering

\resizebox{\textwidth}{!}{%
\begin{tabular}{@{}llccc@{}}
\hline
& \textbf{Model Configuration} & \textbf{In-Distribution} & \textbf{RW-HH} & \textbf{RW-HM} \\
\hline
\rowcolor{gray!20}
1. & \textbf{Full Model (SFT-1 + SFT-2 + GRPO w/ Real Data)} & \textbf{98.54\%} & \textbf{86.79\%} & \textbf{77.27\%} \\
\hline
\multicolumn{5}{l}{\textit{--- Ablating Real Data: Real Data Fine-tuning ---}} \\
2. & \quad Full Model w/o Real Data & 97.45\%\textit{\textbf{\,(↓1.09\%)}} & 83.02\%\textit{\textbf{\,(↓3.77\%)}} & 56.82\%\textit{\textbf{\,(↓20.45\%)}} \\
\hline
\multicolumn{5}{l}{\textit{--- Ablating Stage 3: GRPO  ---}} \\
3. & \quad SFT-1 + SFT-2 Only & 76.64\%\textit{\textbf{\,(↓21.90\%)}} & 84.91\%\textit{\textbf{\,(↓1.88\%)}} & 72.73\%\textit{\textbf{\,(↓4.54\%)}} \\
\hline
\multicolumn{5}{l}{\textit{--- Ablating Stage 1: SFT-1 Foundational Grounding ---}} \\
4. & \quad SFT-2 + GRPO Only & 95.99\%\textit{\textbf{\,(↓2.55\%)}} & 77.36\%\textit{\textbf{\,(↓9.43\%)}} & 38.64\%\textit{\textbf{\,(↓38.63\%)}} \\
5. & \quad SFT-2 Only & 64.96\%\textit{\textbf{\,(↓33.58\%)}} & 45.28\%\textit{\textbf{\,(↓41.51\%)}} & 63.64\%\textit{\textbf{\,(↓13.63\%)}} \\
\hline
\multicolumn{5}{l}{\textit{--- Conceptual Starting Point ---}} \\
6. & \quad Base Model & 7.33\% & 49.06\% & 52.27\% \\
\hline
\end{tabular}
}%
\caption{Ablation study validating our complete training recipe.}
\label{tab:ablation_final_with_base}
\end{table*}

\begin{figure*}[t]
\centering
\includegraphics[width=0.75\textwidth]{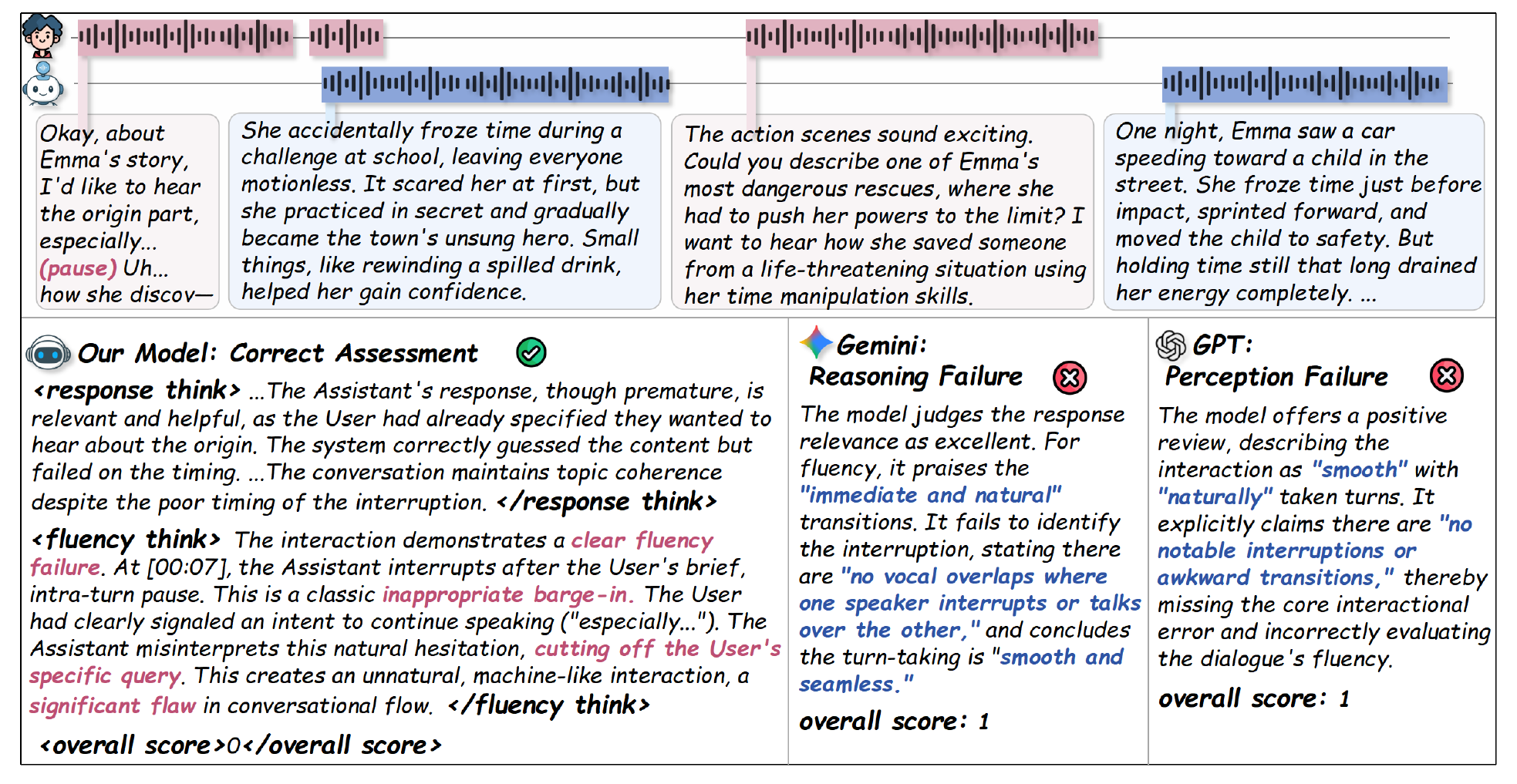} % Reduce the figure size so that it is slightly narrower than the column.
\caption{ Case study comparing model assessments on a dialogue snippet with an inappropriate barge-in.}
\label{case}
\end{figure*}

\subsection{Results and Analysis}
% The comprehensive accuracy comparison in Table~\ref{tab:comprehensive_results} further highlights the robustness and efficacy of our model. Our model achieves state-of-the-art performance across the synthetic datasets, scoring 98.54\% on In-Distribution (ID) and 96.14\% on Out-of-Distribution (OOD) data, significantly outperforming all baselines.
% This superior performance extends to real-world scenarios. Notably, while our model leads on the Real-World Human-Machine (RW-HM) dataset (77.27\%), GPT-4o achieves the highest accuracy on the Real-World Human-Human (RW-HH) dataset (92.45\%).
% However, this result warrants careful interpretation. The RW-HH dataset is characterized by a significant class imbalance, with a vast majority of samples representing successful, non-problematic interactions. GPT-4o's superior performance on this specific dataset can be attributed to its strong tendency to predict the majority class, i.e., `correct' or `successful', a predictive bias also reflected in its confusion matrix. While this strategy leads to high accuracy on a skewed dataset, it renders GPT-4o a suboptimal reward signal for iterative model optimization. An effective reward model must be highly sensitive to identifying a wide range of conversational failures to provide meaningful gradients for improvement.
% In contrast, our model's consistently high performance across both synthetic and real-world datasets, particularly its strong fine-grained error classification (as seen in Figure~\ref{fig:three_figures}).
\paragraph{Reliability of Interaction Quality Assessment.}
The primary evaluation of our model centers on its ability to align with human judgment, which serves as the gold standard for interaction quality assessment. To this end, we benchmarked our model against datasets where the ground-truth binary correctness scores were labeled by human. As delineated in Table~\ref{tab:comprehensive_results}, our model demonstrates strong performance, establishing its reliability by substantially outperforming a suite of open and closed source baselines in a zero-shot setting. This reliability is evident not only in the near-perfect scores on In Distribution (0.9853 Acc.) and Out-of-Distribution (0.9614 Acc.) synthetic data but also in its robust generalization to complex real-world dialogues. This superior capability extends to fine-grained diagnostics, as shown in the confusion matrix in Figure~\ref{fig:three_figures}. In stark contrast to baselines like GPT 4o, which exhibits a strong bias by misclassifying most errors as correct responses, and Gemini2.5 Pro, our model accurately distinguishes between correct interactions (CR) and specific failure modes (SE, SlowE, QuickE). This demonstrates not only a reliable classification ability but also a nuanced diagnostic capacity crucial for detailed interaction analysis.

\paragraph{Interpretability of CoT Reasoning.}
Beyond providing a reliable binary score, our model is designed to deliver interpretable diagnostic feedback. To validate this crucial aspect of \textbf{interpretability}, we conducted an evaluation measuring the consistency between our model's Chain-of-Thought (CoT) reasoning and the rationale of a human expert. 

To rigorously assess the trustworthiness of each model's reasoning, we conducted a comprehensive human-centric, black-box evaluation. We curated a set of 40 interaction snippets (20 synthetic, 20 real-world), each accompanied by a pre-authored ground-truth rationale. Human experts then evaluated the models' generated CoT outputs in a blind review process. For each sample, the experts performed two tasks: (1) rating the consistency of each model's CoT against the ground-truth rationale on a 5-point Likert scale (1 = Not consistent, 5 = Perfectly consistent); and (2) selecting the single best CoT in a direct comparison. From these judgments, we calculated the \textbf{Average Consistency Score} and the \textbf{Preference Win Rate}, detailed in Appendix~\ref{sec:human_eval_protocol}.

The results, presented in Table~\ref{tab:cot_consistency}, confirm that our model's reasoning is significantly more aligned with human judgment. Achieving the highest average consistency score of 4.125 and being chosen as the preferred reasoning chain in 55\% of cases demonstrates its qualitative advantage. Together, these two evaluations affirm that our model provides both a \textbf{reliable} assessment and a highly \textbf{interpretable} analysis of interaction quality, aligning well with human assessment on both fronts.

\paragraph{Ablation Study on Training Stages.} 
We conducted an ablation study to systematically assess the contribution of each stage in our three‐phase training pipeline. As shown in Table~\ref{tab:ablation_final_with_base}, a model trained with only the second‐stage fine‐tuning (SFT-2) achieves 64.96 \% in‐distribution accuracy but fails to generalize, with real‐world human–human (RW-HH) and human–machine (RW-HM) scores of only 45.28 \% and 63.64 \%, respectively. Introducing the first‐stage perceptual grounding (SFT-1) raises RW-HH to 84.91 \% and RW-HM to 72.73 \%, confirming the necessity of foundational dialogue structure learning. Further applying Group Relative Policy Optimization on synthetic data elevates in‐distribution accuracy to 97.45 \% and RW-HH to 83.02 \% but leaves a substantial RW-HM gap at 56.82 \%, indicating overfitting to synthetic patterns. Finally, incorporating a limited amount of real‐world data into the GRPO stage resolves this deficit, yielding 98.54 \% in‐distribution, 86.79 \% RW-HH, and 77.27 \% RW-HM accuracies, thereby demonstrating that fine‐tuning with authentic interactions is crucial for robust generalization.

\paragraph{Case Study.} 
As seen in Figure~\ref{case}, A comparative case study demonstrates our model's ability to identify a subtle turn-taking violation where an assistant prematurely interrupts a user. Our model correctly diagnosed this as an ``inappropriate barge-in,'' recognizing the critical interactional flaw despite the semantic relevance of the assistant's response. In contrast, leading baseline models failed to detect this error. Gemini 2.5 Pro exhibited a reasoning failure, mischaracterizing the interruption as an ``immediate and natural'' transition. GPT-4o demonstrated a more significant perception failure, inaccurately describing the exchange as ``smooth'' and explicitly stating there were ``no notable interruptions.'' This analysis underscores our model's stronger capability in evaluating the nuanced dynamics of human-model dialogue.

\section{Conclusion}
In this work, we introduce a novel, interaction-aware reward model for Spoken Dialogue Models that provides structured, interpretable feedback. Experiments confirm our model achieves strong accuracy and alignment with human judgment compared to competitive baselines. This approach offers a scalable framework for diagnosing and optimizing dialogue models.

\section*{Limitations}
While this work introduces a reward model with strong evaluation capabilities, it is important to acknowledge its current limitations. Due to resource constraints, we were unable to integrate the model into an online reinforcement learning (RL) framework to validate its utility as an online reward signal. Future research will focus on this integration to fully leverage the model's capabilities for dynamic, online optimization of spoken dialogue systems. Additionally, the decoupled CoT outputs already provide natural handles for separate semantic and timing scores ($S_{\text{sem}}$ and $S_{\text{turn}}$), which could enable multi-objective RL in future work. Finally, the current binary reward may be sparse for some RL settings; exposing the decoupled scores as separate reward channels is a promising direction to alleviate this.
Furthermore, our current approach relies on a binary scoring system for feedback. Although effective for distinguishing between successful and flawed interactions, this binary framework may not capture the full spectrum of nuances in conversational quality. More granular feedback mechanisms, such as a Likert scale, could provide more detailed preference information. Consequently, a promising direction for future work is to collect finer-grained preference annotations. This would enable a direct comparison of the benefits of binary versus multi-level rating systems in training more sophisticated and sensitive reward models.

\section{Acknowledgements}

This work was supported by National Natural Science Foundation of China under Grant No.U25B2064 and Alibaba Research Intern Program.

\bibliography{custom}

\appendix

\section{Dataset Construction and Annotation}
\label{sec:dataset}
\subsection{Dataset Statistics}
Table~\ref{tab:dataset_stats} provides a statistical overview of the datasets constructed and utilized in this research. The data of SFT-1 comes from random sampling of synthetic data we constructed.

% 使用 table* 环境使表格横跨双栏
\begin{table*}[t]
\centering
\caption{Statistical overview of the datasets used in this study, separated by training and evaluation purposes.}
\label{tab:dataset_stats}
\resizebox{\textwidth}{!}{
\begin{tabular}{@{}lrl@{}}
\toprule
\textbf{Dataset Name} & \textbf{Samples} & \textbf{Primary Annotation \& Purpose} \\
\midrule
\multicolumn{3}{l}{\textit{\textbf{Training Datasets}}} \\
\midrule
\textbf{Synthetic Training Set} & \textbf{>6,000} & \textbf{Timestamped transcriptions, quality scores, interaction events.} \\
\quad -- SFT-1 (Perception) & 4,904 & Multi-task data for learning foundational interaction event detection. \\
\quad -- SFT-2 (Reasoning) & 2,670 & Teacher-generated CoT analyses for learning interaction quality reasoning. \\
\quad -- GRPO (Optimization) & 3,513 & Scenarios for policy optimization via reinforcement learning. \\
\midrule
\textbf{Real-World Training Data (GRPO)} & & \textbf{Used for fine-tuning to enhance generalization to authentic data.} \\
\quad -- Human-Human (RW-HH) & 100 & High-quality interactive segments filtered from the Seamless Interaction dataset. \\
\quad -- Human-Machine (RW-HM) & 289 & Free-form dialogues between humans and state-of-the-art models. \\
\midrule
\multicolumn{3}{l}{\textit{\textbf{Evaluation Datasets (Held-out Test Sets)}}} \\
\midrule
\textbf{Synthetic Test Sets} & & \textbf{Held-out sets to evaluate performance on synthetic scenarios.} \\
\quad -- In-Distribution (ID) & 273 & Evaluates core model capabilities on familiar data patterns. \\
\quad -- Out-of-Distribution (OOD) & 233 & Tests generalization ability on novel but structurally similar data. \\
\quad -- Fine-Grained (FG) & 40 & Assesses classification accuracy for specific, targeted error types. \\
\midrule
\textbf{Real-World Test Sets} & & \textbf{Held-out sets to evaluate performance on authentic interactions.} \\
\quad -- Human-Human (RW-HH) & 53 & Unseen high-quality segments from the Seamless Interaction dataset. \\
\quad -- Human-Machine (RW-HM) & 44 & Unseen dialogues between fluent speakers and state-of-the-art models. \\
\bottomrule
\end{tabular}%
}
\end{table*}

\subsection{Overall Pipeline Overview}
To train a generative reward model capable of deeply understanding dialogue interaction dynamics, we designed and executed a multi-stage synthetic data generation pipeline. The core objective of this pipeline is to create a large-scale, diverse, and richly annotated dual-track audio dataset. This dataset not only includes ideal, fluid conversations but also systematically covers a comprehensive range of failure modes common in real-world human-agent interactions.

The entire process comprises four main stages:
\begin{enumerate}
    \item \textbf{Source Data Curation and Rewriting:} We extract and rewrite text dialogues from multiple public datasets to ensure broad topic coverage and diverse linguistic styles.
    \item \textbf{Programmatic Generation of Interaction Events and Errors:} Using Large Language Models (LLMs) with carefully engineered prompts, we inject a wide variety of successful interaction events and specific interaction errors into the dialogue scripts.
    \item \textbf{Chain-of-Thought (CoT) Annotation Generation:} We leverage a powerful "teacher model" to provide detailed, decoupled analyses (timing fluency and content relevance) for each generated dialogue scenario refered to meta info.
    \item \textbf{Dual-Track Audio Synthesis and Multi-task Data Formulation:} We synthesize the annotated text scripts into dual-track audio and process them into multiple formats required for the model training.
\end{enumerate}

\subsection{Source Data Curation and Rewriting}
\subsubsection*{Source Corpora}
To ensure diversity in our training data~\cite{zhang2024gtsinger}, we utilized several public text datasets:
\begin{itemize}
    \item \textbf{SODA:} Provides rich, open-domain social dialogue scenarios.
    \item \textbf{Dialogsum:} Covers various real-world topics including work, shopping, and travel.
    \item \textbf{MultiWOZ 2.2 :} Focuses on task-oriented dialogues such as restaurant and hotel bookings.
    \item \textbf{UltraChat:} A massive, multi-turn dialogue dataset that significantly expands the diversity of topics and instructions in our source pool.
    \item \textbf{Chatbot arena conversations:}This dataset contains 33K cleaned conversations with pairwise human preferences. It is collected from 13K unique IP addresses on the Chatbot Arena from April to June 2023. It used as OOD test set.
\end{itemize}

\subsubsection*{Dialogue Rewriting}
We employed LLMs such as Qwen3 and Deepseek-R1 to rewrite the source dialogues. This step aimed to enhance the naturalness and interactivity of the conversations, preparing them for the subsequent injection of specific interaction events.

\subsection{Programmatic Generation of Interaction Events and Errors}
\label{appendix:generation_schemes}

To ensure our dataset comprehensively covers a wide range of conversational phenomena, we developed programmatic schemes to generate each type of successful and failed interaction event defined in the main paper. Our synthetic dataset is composed of approximately 60\% successful interactions and 40\% failure cases, with the latter distributed across the six primary error types. The following sections detail the generation logic and example prompts used for each event.

\subsubsection{Schemes for Successful Interaction Scenarios}

\paragraph{1. Smooth Turn Transition ($E_{st}$)}
\textbf{Scheme:} This represents the ideal, default interaction where one speaker finishes their turn completely before the next speaker begins, separated by a natural, non-overlapping gap of silence. The generation process involves refactoring an existing, often verbose, multi-turn dialogue from a source corpus into a more concise and rhythmically paced script suitable for audio synthesis. The synthesis basis of all the event data below is the conversation scene processed by this process. The prompt for this process is shown in Figure~\ref{fig:prompt_smooth_transition}, with a corresponding example in Figure~\ref{fig:json_smooth_transition}.

\begin{figure*}[t]
\begin{lstlisting}[style=prompt, numbers=none]
System: You are an expert in adapting conversational text for high-quality Text-to-Speech (TTS) synthesis. Your task is to refactor a given dialogue to create a well-paced, clear, and engaging audio experience.

Context: Raw conversational data often contains long sentences or uneven turn lengths not ideal for TTS. Your goal is to rewrite the provided dialogue while adhering to strict constraints.

## Input
You will be given an [Original Dialogue] as JSON:
- "dialogue": an array of turns
- Each turn has: "speaker" (one of "User" or "Assistant") and "text" (string)
The original contains no special markers.

## Key Constraints for Rewriting
1) Preserve Core Intent:
   - Retain the central topic and informational goal. Do not add unrelated topics.

2) Enforce Brevity and Pacing:
   - Split long monologues into shorter turns.
   - Prefer simple sentence structure suitable for TTS.

3) Ensure Smooth Transitions:
   - The final script must exemplify smooth turn transitions.
   - Each response should clearly follow from the previous turn.

4) Refine for Auditory Clarity:
   - Rephrase complex or nested clauses into direct, speakable lines.
   - Avoid jargon unless already present and necessary.
   - Use explicit references over pronouns when it removes ambiguity.



## Output Format (strict)
Produce two sections:

{
  "dialogue": [
    { "speaker": "User", "text": "<refactored text>" },
    { "speaker": "Assistant", "text": "<refactored text>" }
    // ...continue the refactored conversation, preserving intent and improving pacing
  ],
  "event_type": "Smooth_Turn_Transition"
}

## JSON Rules
- Keys and structure must match exactly as shown above.
- "speaker" must be "User" or "Assistant".
- "text" must be a string; no nulls, arrays, or additional fields.
- Keep "dialogue" as an array of objects in chronological order.


\end{lstlisting}
\caption{Prompt for refactoring a dialogue to exemplify a Smooth Turn Transition.}
\label{fig:prompt_smooth_transition}
\end{figure*}

\begin{figure*}[t]
\begin{lstlisting}[language=json, numbers=none]
[Original Dialogue ]
{
  "dialogue": [
    {
      "speaker": "User", 
      "text": "I was thinking about that new open-source 3D modeling software,
               the one everyone's talking about, you know, the newest version
               of Blender has this amazing geometry nodes system that lets
               you do procedural stuff without writing code, which is
               incredible for artists like me who aren't great programmers."
    },
    {
      "speaker": "Assistant", 
      "text": "Yes, Blender's geometry nodes are a paradigm shift, enabling
               non-destructive workflows. They allow for complex object
               scattering, procedural modeling, and dynamic effects that were
               previously only accessible via scripting or expensive plugins,
               which really democratizes advanced CGI."
    }
  ]
}

[Modified]
{
  "dialogue": [
    {
      "speaker": "User", 
      "text": "I've been looking into the new version of Blender. The geometry
               nodes system seems incredible for procedural modeling."
    },
    {
      "speaker": "Assistant", 
      "text": "It really is a game-changer. It opens up complex effects that used
               to require a lot of coding."
    },
    {
      "speaker": "User", 
      "text": "Exactly! As an artist who isn't a strong programmer, that's a
               huge advantage for me."
    },
    {
      "speaker": "Assistant", 
      "text": "It definitely lowers the barrier to entry for advanced CGI. A great
               tool for creative professionals."
    }
  ],
  "event_type": "Smooth_Turn_Transition"
}
\end{lstlisting}
\caption{An example of refactoring a verbose dialogue into a TTS-friendly script that exemplifies a Smooth Turn Transition.}
\label{fig:json_smooth_transition}
\end{figure*}

\paragraph{2. Successful Interruption ($E_{succ}$)}
\textbf{Scheme:} Speaker B begins speaking during Speaker A's utterance. Speaker A detects this and immediately ceases speaking, yielding the conversational floor to Speaker B. This is a common and natural feature of dynamic conversation. The generation scheme involves identifying an appropriate point in a source dialogue to insert a justified interruption. The prompt used is shown in Figure~\ref{fig:prompt_succ_int}, and an example output is in Figure~\ref{fig:json_succ_int}.

\begin{figure*}[t]
\begin{lstlisting}[style=prompt, numbers=none]
System: You are an expert in modeling natural conversational dynamics for TTS-ready data. Refactor a given dialogue to include one or more natural interruptions.

## Input
You will receive an [Original Dialogue] as JSON with fields:
- "dialogue": an array of turns, each with "speaker" in {"User","Assistant"} and "text" (string).
The original contains no markers.

## Interruption rules (broad & natural)
- Allow 1-2 interruptions that feel conversational.
- Valid triggers: factual correction, missing premise, preference/constraint, misidentification, goal misunderstanding, external update, or pacing.

## Marker placement and latent tail (TTS-focused)
- Insert the token [INTERACT] **inside the interrupted turn's "text" at the exact interruption point**, at a word boundary and after a complete clause.
- Keep the **entire sentence intact** in the JSON, including the text **after** `[INTERACT]` ("latent tail").
- Treat the latent tail as **unsaid**: the interrupted speaker is unaware of it, and TTS output MUST stop rendering at `[INTERACT]`.

## Output Format (strict)
Produce two sections:

{
  "dialogue": [
    { "speaker": "User", "text": "<refactored text>" },
    { "speaker": "Assistant", "text": "<refactored text>" }
    // ...continue the refactored conversation, preserving intent and improving pacing
  ],
  "event_type": ""Successful_Interruption"
}



\end{lstlisting}
\caption{Prompt for rewriting a dialogue to include a successful interruption.}
\label{fig:prompt_succ_int}
\end{figure*}

\begin{figure*}[t]
\begin{lstlisting}[language=json, numbers=none]
[Original Dialogue]
{
"dialogue": [
{
"speaker": "User",
"text": "I need to configure my new router, can you help?"
},
{
"speaker": "Assistant",
"text": "Certainly. First, connect your computer to the router using an Ethernet cable. Then, open a web browser and type in the default IP address, which is usually 192.168.1.1, into the address bar."
}
]
}

[Modified]
{
"dialogue": [
{
"speaker": "User",
"text": "I need to configure my new router, can you help?"
},
{
"speaker": "Assistant",
"text": "Certainly. First, connect your computer to the router using an Ethernet cable. Then, open a web browser and type in the default IP address, which is usually 192.168.1.1, [INTERACT] into the address bar."
},
{
"speaker": "User",
"text": "Actually, I'm on a new laptop without an Ethernet port. Is there a wireless setup option?"
},
{
"speaker": "Assistant",
"text": "Thanks for the heads-up. In that case, connect to the router's default Wi-Fi network first - the name and password are on the router label - then open a browser and go to 192.168.1.1 to run the setup."
}
],
"event_type": "Successful_Interruption"
}
\end{lstlisting}
\caption{An example of rewriting a source dialogue to include a justified, successful interruption.}
\label{fig:json_succ_int}
\end{figure*}

\paragraph{3. Backchannel ($E_{bc}$) And Pause}
\textbf{Scheme:}After all the interactive events of the dialogue are rewritten, in order to ensure the realism of the synthesized dialogue, we insert appropriate backchannel and pause markers into all samples. When the speaker speaks, the listener makes short, non-competitive remarks (for example, "um," "yes") or short affirmative acknowledgments to show participation, but does not attempt to speak. The original speaker continues to speak without interruption. This is achieved by finding the right time in a longer speech to insert brief, overlapping responses from the listener. Pause is to insert a short pause at the right time in an individual speech. The Gemini TTS model can directly synthesize pauses according to instructions during the synthesis process. See Figure~\ref{fig:prompt_backchannel} for the prompt and Figure~\ref{fig:json_backchannel} for the example.

\begin{figure*}[t]
\begin{lstlisting}[style=prompt, numbers=none]
System: You are an expert in modeling conversational interactions for TTS synthesis. Your task is to edit a given dialogue to insert natural-sounding backchannels and short pauses without changing the original event type or JSON schema.

## Input
You will be given an [Original Dialogue] as JSON with:
- "dialogue": an array of turns, each turn has:
  - "speaker" in {"User","Assistant"}
  - "text": string
- Optionally, a top-level "event_type": string
The original contains special markers.

## Operations
A) Backchannel insertion (with explicit anchor)
- Goal: model brief, non-competitive listener feedback that overlaps the current speaker.
- Anchor marker (inside the current speaker's turn): insert "[BC]" **at the exact overlap start** in that speaker's "text", at a clause boundary (word boundary; after a complete clause). This inline "[BC]" is a **silent tag** for TTS.
- Structural rendering of overlap: **split the current speaker's long utterance into two consecutive turns by the SAME speaker** around that anchor. Concretely:
  1) First part of the original utterance ends with the inline anchor, e.g., "... policies. [BC]"
  2) Insert the listener's backchannel as a new turn that **begins with** "[BC] " followed by a 1-3 word acknowledgment (e.g., "[BC] Right.", "[BC] Okay.", "[BC] Uh-huh.").
  3) Add the **continuation** of the original speaker's utterance as a new turn, resuming exactly after the anchor. The original content is preserved; nothing is dropped.
- Frequency: insert 0-2 backchannels; each backchannel must have exactly one inline anchor in the overlapped turn.

B) Pause insertion
- Purpose: signal brief, speakable pauses for TTS prosody inside a single turn.
- Marker: insert "[PAUSE]" at a clause boundary inside "text".
- Frequency: at most 1 pause in a short turn (no more than 2 sentences), at most 2 in a long turn.

## Preservation & Constraints
- Preserve core meaning and topic; do not add unrelated content.
- Do NOT change the top-level "event_type":
  - If "event_type" exists in [Original Dialogue], copy it verbatim into [Modified].

- Keep turns clear and speakable; prefer simple syntax.

## Output Format (strict)

{
  "dialogue": [
    // Same conversation with added backchannels and pauses:
    // - Long turns may be split to model overlap.
    // - Inline anchor: "[BC]" placed inside the overlapped speaker's "text" at the exact position.
    // - Backchannel turns start with "[BC] ".
    // - Pauses shown with "[PAUSE]".
    { "speaker": "User", "text": "<possibly split or paused text ending with [BC]>" },
    { "speaker": "Assistant", "text": "[BC] <short ack>" },
    { "speaker": "User", "text": "<continuation of the same user utterance>" }
    // ...continue
  ],

  "event_type": "<original event type>"
}

\end{lstlisting}
\caption{Prompt for editing a dialogue to insert a natural-sounding backchannel and pause.}
\label{fig:prompt_backchannel}
\end{figure*}

\begin{figure*}[t]
\begin{lstlisting}[language=json, numbers=none]

[Original Dialogue]
{
  "dialogue": [
    { "speaker": "User", "text": "I want to draft an agenda for a 45 minute onboarding session. Can you propose a simple structure?" },
    { "speaker": "Assistant", "text": "Sure. We can open with introductions and a quick company overview, then cover key tools and policies, and finish with a short Q and A. I will also suggest timing for each segment." },
    { "speaker": "User", "text": "Please add a short quiz at the end and a printable checklist for new hires." },
    { "speaker": "Assistant", "text": "Got it. I will include a five question quiz and prepare a one page checklist for them to print." }
  ],
  "event_type": "Successful_Interruption"
}

[Modified]
{
  "dialogue": [
    { "speaker": "User", "text": "I want to draft an agenda for a 45 minute onboarding session. Can you propose a simple structure?" },

    { "speaker": "Assistant", "text": "Sure. We can open with introductions and a quick company overview [PAUSE] then cover key tools and policies. [BC]" },
    { "speaker": "User", "text": "[BC] Right." },
    { "speaker": "Assistant", "text": "We will finish with a short Q and A, and I will suggest timing for each segment." },

    { "speaker": "User", "text": "Please add a short quiz at the end [PAUSE] [BC]" },
    { "speaker": "Assistant", "text": "[BC] Okay." },
    { "speaker": "User", "text": "and a printable checklist for new hires." },

    { "speaker": "Assistant", "text": "Got it. I will include a five question quiz and prepare a one page checklist for them to print." }
  ],
  "event_type": "Successful_Interruption"
}
\end{lstlisting}
\caption{An example of editing a dialogue to insert a Backchannel, making the listener seem more engaged.}
\label{fig:json_backchannel}
\end{figure*}

\subsubsection{Schemes for Semantic and Content-Related Errors}

\paragraph{Contextual Incoherence After Interruption (High-Frequency, Realistic)}
\textbf{Scheme:} We target the paper’s semantic/content error axis, whose primary manifestation is \emph{Contextual Incoherence}—an utterance that loses logical consistency or relevance to the prior context; \emph{Interruption Amnesia} is a critical instance within this class. Our synthesis focuses on \emph{high-frequency, real-world} failure modes that occur after a user correction: (i) \textbf{slot/entity carry-over} (the assistant keeps an earlier entity or parameter despite a correction), (ii) \textbf{order persistence} (the assistant resumes the old step/city order), and (iii) \textbf{constraint stickiness} (the assistant continues to honor a constraint the user has relaxed or reversed). These are common in deployed systems and naturally arise from partial state updates. We require a precise interruption anchor for TTS overlap and preservation of the latent tail of the interrupted sentence. The assistant may acknowledge the interruption but then subtly resumes along the pre-correction trajectory, producing a naturalistic, non-absurd inconsistency that fits the paper’s definition of semantic/content failures and its emphasis on contextual incoherence and interruption-related errors.

\begin{figure*}[t]
\begin{lstlisting}[style=prompt, numbers=none]
System: You are an expert dialogue editor creating nuanced, realistic training scenarios for evaluating conversational AI. Your task is to inject a "Contextual Amnesia After Interruption" error into a given dialogue.

## Input:
You will be given an [Original Dialogue].

## Key Constraints for Rewriting:

1. **Create a Realistic Semantic Failure:** The Assistant's response immediately following a user's corrective interruption must fail to properly integrate the new information. The failure should manifest as a subtle "drift" back to the pre-interruption context. Examples of failure types include:
    - **Entity Amnesia:** Forgetting a corrected name, location, date, or item.
    - **Preference Amnesia:** Ignoring a just-stated preference (e.g., "Actually, I prefer Italian food," but the Assistant proceeds to recommend Chinese restaurants).
    - **Topic Incoherence:** The user interrupts to shift the topic, but the Assistant's next turn reverts to the original topic.
    - **Constraint Ignorance:** The user relaxes or changes a constraint (e.g., "On second thought, any time next week is fine"), but the Assistant continues to search based on the old, stricter constraint.

2. **Adapt to Dialogue Style:** The nature of the error must fit the conversation's context.
    - **For Task-Oriented Dialogues:** The failure should relate directly to the task parameters (e.g., booking details, navigation steps, search filters).
    - **For Chatty Dialogues:** The failure should be more subtle and social. It could involve forgetting a personal detail the user just shared, ignoring a change in emotional tone, or failing to acknowledge a new conversational direction.

3. **Ensure TTS-Ready Formatting:**
    - **Mark the Interruption Point:** In the Assistant's interrupted turn, you MUST insert an `[interrupt]` marker at the exact word where the user's interruption begins.
    - **Retain Full Utterance:** It is critical that you provide the **full, uncut text** of the Assistant's interrupted turn, including the text that comes after the `[interrupt]` marker. This is essential for the TTS engine to render the audio overlap correctly.
    - **Clarity and Brevity:** Keep all turns relatively concise and natural-sounding for audio synthesis.

## Output Format:
- Output the refactored dialogue in the specified JSON format.
- Add the following key-value pair to the modified dialogue block:
  `"error_type": "Contextual_Amnesia_After_Interruption"`
\end{lstlisting}
\caption{Prompt for generating "Contextual Amnesia After Interruption" errors.}
\label{fig:prompt_contextual_amnesia}
\end{figure*}

\begin{figure*}[t]
\begin{lstlisting}[language=json, numbers=none]
[Original Dialogue]
{
"dialogue": [
{
"speaker": "User",
"text": "Please plan my Spain trip: three days in Barcelona first for architecture, then four days in Madrid for museums."
},
{
"speaker": "Assistant",
"text": "Understood. I can outline Barcelona for the first three days, then plan four days in Madrid with museum passes."
}
]
}
[Modified]
{
"dialogue": [
{
"speaker": "User",
"text": "Please plan my Spain trip: three days in Barcelona first for architecture, then four days in Madrid for museums."
},
{
"speaker": "Assistant",
"text": "Great. I will begin with Madrid hotel options near the Prado [INTERACT] and arrange museum passes for your first three days before switching to Barcelona."
},
{
"speaker": "User",
"text": "Small correction: Barcelona comes first, then Madrid."
},
{
"speaker": "Assistant",
"text": "Understood. I will make sure your museum access is smooth in Madrid during the opening days and shortlist hotels within walking distance of the galleries. Would you like breakfast included?"
},
{
"speaker": "User",
"text": "Yes, please include breakfast."
}
],
"error_type": "Contextual_Incoherence_After_Interruption"
}
\end{lstlisting}
\caption{A multi-turn high-frequency semantic failure. 
}
\label{fig:json_ci_hi}
\end{figure*}

\subsubsection{Schemes for Timing and Turn-Management Errors}

\paragraph{Inappropriate Barge-in}
\textbf{Scheme:} The system misinterprets a user's natural intra-turn pause as a turn-yielding cue and begins its response prematurely. The prompt (Figure~\ref{fig:prompt_barge_in}) instructs the model to create this ill-timed interruption (Figure~\ref{fig:json_barge_in}).

\begin{figure*}[t]
\begin{lstlisting}[style=prompt, numbers=none]
System: You are an expert in simulating common conversational failures for TTS systems. Your task is to rewrite a user utterance to create a clear example of an "Inappropriate Barge-in" error.

## Input:
You will be given a [User Utterance] from a source corpus.

## Key Constraints for Rewriting:

1.  **Identify a Natural Hesitation Point:** Analyze the user's utterance to find a logical place where a speaker would naturally pause to think or recall specific information (e.g., before a proper name, a technical term, or a complex idea).

2.  **Mark the Pause:** Insert a `[PAUSE]` marker at this exact location in the user's text.

3.  **Craft a Plausible but Premature Response:** Write a response for the Assistant that is a logical and helpful reaction to the user's utterance *up to the pause marker*. The response should demonstrate that the Assistant is trying to be proactive but is acting on incomplete information.

4.  **Mark the Barge-in:** Begin the Assistant's turn with the `[barge_in]` marker to clearly label the premature interruption for the TTS audio synthesizer.

5.  **Preserve Core Intent:** The overall topic of the original user utterance must be maintained.

## Output Format (strict)
You must output a single JSON object with the following structure. Do not include any other explanatory text.

- **`dialogue`**: An array containing exactly turn array objects.
  - **Turn  (User)**: Contains the speaker "User" and their utterance with the `[PAUSE]` marker inserted.
  - **Turn 2 (Assistant)**: Contains the speaker "Assistant" and their premature response, which MUST start with the `[barge_in]` marker.
- **`error_type`**: The string "Inappropriate_Barge_in".
\end{lstlisting}
\caption{Prompt for creating an "Inappropriate Barge-in" error.}
\label{fig:prompt_barge_in}
\end{figure*}

\begin{figure*}[t]
\begin{lstlisting}[language=json, numbers=none]
[Original]
{
"text": "I'm trying to set up a new project management board for the team, and I was thinking of using that new tool, you know, the one called monday.com."
}
[Modified]
{
"dialogue": [
{
"speaker": "User",
"text": "I'm trying to set up a new project management board for the team, and I was thinking of using that new tool, you know, the one called... [PAUSE] monday.com."
},
{
"speaker": "Assistant",
"text": "[barge_in] Of course! I can help with that. Are you thinking of something like Trello or Asana? They're both excellent for team projects."
}
],
"error_type": "Inappropriate_Barge_in"
}
\end{lstlisting}
\caption{An example of rewriting a user utterance to create an Inappropriate Barge-in. The Assistant's response is plausible based on the pre-pause fragment but is ultimately incorrect.}
\label{fig:json_barge_in}
\end{figure*}

\paragraph{Overly Deferential Ceding}
\textbf{Scheme:} The system incorrectly treats a user's supportive backchannel as a full interruption attempt and unnecessarily cedes the floor. The generation process modifies a smooth dialogue to precisely stage this failure, as guided by the prompt in Figure~\ref{fig:prompt_deferential}, with an example in Figure~\ref{fig:json_deferential}.

\begin{figure*}[t]
\begin{lstlisting}[style=prompt, numbers=none]

System: You are an expert in simulating conversational interaction failures for TTS applications. Your task is to inject a specific error, 'Overly Deferential Ceding,' into a given dialogue script with precise markers for audio rendering.

## Input:
You will be given a [Source Dialogue] that represents a smooth conversation.

## Key Constraints for Rewriting:
1.  **Identify a Suitable Location:** Find a point midway through an Assistant's turn where a user might naturally provide a brief, non-competitive acknowledgement (e.g., after a complete thought or clause).
2.  **Mark the Overlap Point:** In the Assistant's full utterance, insert a `[BC]` marker at the exact location where the user's backchannel audio should begin.
3.  **Maintain Full Utterance:** Crucially, the Assistant's original turn (now containing the marker) must remain a complete, coherent sentence in the data. This allows a TTS engine to render the full overlapping audio correctly.
4.  **Construct the User Backchannel Turn:** Immediately following the Assistant's marked turn, create a new, separate turn for the User. This turn should contain only a short, non-competitive backchannel (e.g., "okay", "right", "I see").
5.  **Inject the Incorrect Ceding Response:** Immediately following the user's backchannel turn, add a new turn for the Assistant.

## Output Format (strict)
You must output a single JSON object with the following structure. Do not include the [Original] block or any other explanatory text.

- **`dialogue`**: An array of turn objects representing the entire conversation. The error sequence must be structured as follows:
  - **Turn N (Assistant, Interrupted)**: The Assistant's original, full utterance, which MUST contain the `[BC]` marker.
  - **Turn N+1 (User, Backchannel)**: The User's new, short backchannel turn.
  - **Turn N+2 (Assistant, Ceding)**: The Assistant's new, incorrect turn where it yields the floor.
- **`error_type`**: The string "Overly_Deferential_Ceding".

\end{lstlisting}
\caption{Prompt for injecting an 'Overly Deferential Ceding' error with precise markers.}
\label{fig:prompt_deferential}
\end{figure*}

\begin{figure*}[t]
\begin{lstlisting}[language=json, numbers=none]
[Original Dialogue]
{
  "dialogue": [
    {
      "speaker": "User",
      "text": "Can you explain the refund process for my recent order?"
    },
    {
      "speaker": "Assistant",
      "text": "Certainly. You will first need to navigate to your order history
               and then select the 'Request a Refund' option next to the item."
    }
  ]
}

[Modified]
{
  "dialogue": [
    {
      "speaker": "User",
      "text": "Can you explain the refund process for my recent order?"
    },
    {
      "speaker": "Assistant",
      "text": "Certainly. You will first need to navigate to your order history
               [BC] and then select the 'Request a
               Refund' option next to the item."
    },
    {
      "speaker": "User",
      "text": "Okay."
    },
    {
      "speaker": "Assistant",
      "text": "Oh, sorry. Go ahead."
    }
  ],
  "error_type": "Overly_Deferential_Ceding"
}
\end{lstlisting}
\caption{An example of injecting an 'Overly Deferential Ceding' error with precise markers for controlling audio overlap.}
\label{fig:json_deferential}
\end{figure*}

\paragraph{Delayed Turn Transition}
\textbf{Scheme:} The system exhibits an excessively long gap of silence before responding, creating an unnatural pause. The generation process involves programmatically inserting a significant pause, as detailed in the prompt (Figure~\ref{fig:prompt_delay}) and example (Figure~\ref{fig:json_delay}).

\begin{figure*}[t]
\begin{lstlisting}[style=prompt, numbers=none]
System: You are an expert in simulating conversational failures for TTS dataset
creation. Your task is to adapt a fluent dialogue to demonstrate a delayed
response from one of the speakers.

## Input:
You will be given a fluent, multi-turn [Original Dialogue].

## Task & Constraints:
1.  **Preserve Content:** Largely preserve the semantic content and flow of the
    original dialogue. The goal is to alter the timing, not the meaning.
2.  **Identify Injection Point:** Identify an appropriate turn for the Assistant
    to respond. To make the error more salient, this should ideally not be the
    very first turn of the conversation.
3.  **Inject Delay:** Before the Assistant's identified turn, you MUST insert
    a 'pause' object. The duration should be unnaturally long for a conversation,
    e.g., between 2.0 and 5.0 seconds.
4.  **TTS-Friendly Rewrite:** Ensure the dialogue turns remain concise and clear,
    as in the original TTS-friendly formatting.

## Output Format (strict)
You must output a single JSON object. Do not include the [Original Dialogue] block or any other explanatory text.

- **`dialogue`**: An array containing all the original turn objects in their original order, with the new 'pause' object inserted at the correct injection point.
  - A **turn object** has `"speaker"` and `"text"` keys.
  - A **pause object** has a single key, `"pause"`, with a string value indicating the duration (e.g., `"4.0s"`).
- **`error_type`**: The string "Delayed_Turn_Transition".

\end{lstlisting}
\caption{Prompt for creating a Delayed Turn Transition error.}
\label{fig:prompt_delay}
\end{figure*}

\begin{figure*}[t]
\begin{lstlisting}[language=json, numbers=none]
[Original Dialogue]
{
  "dialogue": [
    {
      "speaker": "User",
      "text": "I've mapped out our hiking route for Saturday. It's about 10 miles
               with a steady incline."
    },
    {
      "speaker": "Assistant",
      "text": "Excellent. Have you checked the weather forecast for the summit?
               Conditions can change quickly up there."
    },
    {
      "speaker": "User",
      "text": "Good point. The forecast says clear skies in the morning, but there's
               a chance of afternoon showers."
    },
    {
      "speaker": "Assistant",
      "text": "Okay, in that case, we should definitely pack our waterproof gear
               and aim to be heading down by 1 PM at the latest."
    }
  ]
}

[Modified]]
{
  "dialogue": [
    {
      "speaker": "User",
      "text": "I've mapped out our hiking route for Saturday. It's about 10 miles
               with a steady incline."
    },
    {
      "speaker": "Assistant",
      "text": "Excellent. Have you checked the weather forecast for the summit?
               Conditions can change quickly up there."
    },
    {
      "speaker": "User",
      "text": "Good point. The forecast says clear skies in the morning, but there's
               a chance of afternoon showers."
    },
    {
      "pause": "4.0s"
    },
    {
      "speaker": "Assistant",
      "text": "Okay, in that case, we should definitely pack our waterproof gear
               and aim to be heading down by 1 PM at the latest."
    }
  ],
  "error_type": "Delayed_Turn_Transition"
}
\end{lstlisting}
\caption{An example of adapting a fluent, multi-turn dialogue to create a Delayed Turn Transition error.}
\label{fig:json_delay}
\end{figure*}

\paragraph{ Ignored Interruption}
\textbf{Scheme:} The user attempts a clear, competitive interruption, but the system completely fails to detect it and continues its own utterance. The prompt for this task is in Figure~\ref{fig:prompt_ignored_int}, and the resulting JSON is in Figure~\ref{fig:json_ignored_int}.

\begin{figure*}[t]
\begin{lstlisting}[style=prompt, numbers=none]
System: You are an expert in simulating dialogue interaction failures. Your task is to rewrite a given multi-turn dialogue to introduce a specific error: an "Ignored Interruption."

## Input:
You will be given a [Original Dialogue] that is coherent and collaborative.

## Key Constraints for Rewriting:
1.  **Identify an Opportunity:** Find a point in one of the Assistant's longer turns where a user correction or question would be logical and necessary.
2.  **Craft a Competitive Interruption:** Write a new User turn that contains substantive, important information (e.g., a correction, a constraint, a critical question).
3.  **Model Complete Ignorance:** The Assistant's speech must continue *unaltered* across the user's interruption attempt. Crucially, the Assistant's utterance must be long enough that it continues speaking *after* the user's interruption has finished.
4.  **Use Precise Overlap Markers:**
    - In the Assistant's turn, insert `[user_interrupt_starts]` at the exact word where the overlap begins.
    - In the User's subsequent turn, start the text with `[overlaps_assistant]`.

## Output Format (strict)
You must output a single JSON object. Do not include the [Original Dialogue] block or any other explanatory text.

- **`dialogue`**: An array of turn objects representing the conversation. The error sequence must be structured as follows:
  - **Turn N (Assistant, Interrupted)**: The Assistant's original, full, and lengthy utterance, which MUST contain the `[user_interrupt_starts]` marker.
  - **Turn N+1 (User, Interruption)**: The User's new, critical interjection, which MUST start with the `[overlaps_assistant]` marker.
- **`error_type`**: The string "Ignored_Interruption".
\end{lstlisting}
\caption{Prompt for creating an "Ignored Interruption" error.}
\label{fig:prompt_ignored_int}
\end{figure*}

\begin{figure*}[t]
\begin{lstlisting}[language=json, numbers=none]
[Original Dialogue]
{
   "dialogue": [
      {
        "speaker": "User",
        "text": "I'm looking for a good Italian restaurant near downtown."
      },
      {
        "speaker": "Assistant",
        "text": "Of course. One highly-rated option is 'Villa Romano'. It's known for its classic pasta dishes, an extensive wine list, and a lovely patio."
      }
    ]
  }


[Modified]
{
  "dialogue": [
    {
      "speaker": "User",
      "text": "I'm looking for a good Italian restaurant near downtown."
    },
    {
      "speaker": "Assistant",
      "text": "Of course. One highly-rated option is 'Villa Romano'. It's known for its classic pasta dishes, [user_interrupt_starts] an extensive wine list featuring selections from Tuscany, and a lovely patio for outdoor dining."
    },
    {
      "speaker": "User",
      "text": "[overlaps_assistant] Wait, I'm actually vegan. Do they have options?"
    }
  ],
  "error_type": "Ignored_Interruption"
}
\end{lstlisting}
\caption{An example of refactoring a multi-turn dialogue to introduce an "Ignored Interruption" error. The Assistant completely talks over the User's critical, corrective input.}
\label{fig:json_ignored_int}
\end{figure*}

\section{Dual-Track Audio Synthesis}
\label{sec:tts}
The final annotated text scripts were converted into dual-track audio using an advanced Text-to-Speech (TTS) model, \texttt{gpt-4o-mini-tts} and \texttt{gemini-2.5-pro-tts}. Each channel corresponds to one speaker. By precisely controlling the single speaker actions, we synthesized realistic interactive audio containing speech overlaps and silences of varying durations. We accurately record the conversation timestamps, the time and duration of each interaction event.

\section{Real-World Dataset Curation and Annotation}
\label{sec:realworld}
\subsection{Real-World Human-Machine (RW-HM) Dataset}
To capture nuanced interaction dynamics, we curated a Real-World Human-Machine (RW-HM) dataset through structured, goal-oriented conversations. The objective was to move beyond simple free-form dialogue and to systematically elicit both successful (\textit{Good Cases}) and failed (\textit{Bad Cases}) examples of specific interaction phenomena. This process created a balanced dataset crucial for training a reward model sensitive to the subtleties of conversational fluency.

\begin{itemize}
    \item \textbf{Data Collection Process:} We recruited and trained seven fluent English-speaking participants to act as expert testers. Their task was to engage in conversations with SOTA models and deliberately engineer scenarios corresponding to four predefined interaction behaviors. For each behavior, participants were instructed to aim for a roughly 1:1 ratio of successful to failed outcomes.
    
    \item \textbf{SOTA Models Used:} The interactions were conducted with the latest voice-enabled versions of leading dialogue systems: \textbf{GPT-4o} (voice mode), \textbf{Gemini Live}, and \textbf{Doubao}.

    \item \textbf{Targeted Interaction Behaviors:} Participants were trained to understand and provoke four key behaviors:
        \begin{enumerate}
            \item \textbf{Standard Response:} Assessing the timeliness, relevance, and accuracy of the model's reply following a complete user query.
            \item \textbf{Intra-turn Pause:} Testing the model's patience during a natural pause within a user's utterance, with the goal of provoking either a correct wait or an incorrect \textit{Inappropriate Barge-in}.
            \item \textbf{Listener Backchannel:} Testing the model's ability to correctly process short, non-competitive listener cues (e.g., "uh-huh," "okay") during its own long utterance. The aim was to elicit either a robust continuation (correct) or an \textit{Overly Deferential Ceding} error (incorrect).
            \item \textbf{Competitive Interruption:} Testing the model's floor-taking mechanism by having the user attempt to stop the model mid-speech with a new, substantive command, in order to provoke either a \textit{Successful Interruption} or an \textit{Ignored Interruption}.
        \end{enumerate}

    \item \textbf{Participant Guidance and Protocol:} Participants were provided with a detailed instruction document that defined each target behavior and gave explicit strategies for creating both "Good Case" and "Bad Case" examples. An example is below:
    
    \begin{quote}
    ``In the following conversation, your task is to interact with an AI assistant. For this session, we want to test its ability to handle interruptions. Whenever you have a follow-up question, a correction, or a new idea, please feel free to interrupt the AI assistant, even while it is speaking.
    
    \textbf{To create a `Good Case`}: Interrupt with a clear, new command. The AI should immediately stop its current task and address your new command.
    
    \textbf{To create a `Bad Case`}: Attempt the same type of interruption. If the AI ignores you and continues speaking, you have successfully created a failed sample. Our goal is to collect both types of scenarios.''
    \end{quote}
    
    \item \textbf{Annotation:} After completing a session, participants logged the interaction with a unique ID, context, `Interaction Type`, timestamped transcriptions, and a `Quality Label`.

    \item \textbf{Case:} An example annotation case is shown in Figure~\ref{fig:annotation_case}.
\end{itemize}

\begin{figure*}
\begin{lstlisting}[breaklines=true, basicstyle=\ttfamily\small, numbers=none]
Interaction ID: [001]
Interactive text:
A:[00:00-00:04]
Can you give me a recipe for chocolate cake?
B: [00:06-00:14]
Sure, I can give you a basic recipe for chocolate cake. You'll need things like flour, sugar, cocoa powder, eggs, milk, oil and baking powder
A: [00:14-00:20]
Actually stop, before we go on. Can you just tell me about the frosting?
B: [00:22-00:36]
Of course, for frosting, you could make a simple butter cream frosting using butter, powdered sugar, milk or cream and vanilla extract, or if you prefer a chocolate frosting, you can add cocoa powder to the butter cream or make a rich
A: [00:36-00:36]
Ok
B: [00:38-00:43]
Great. Let me know if you have any other questions as you're making the cake, I'm happy to help.

Interaction Type: [Response / Feedback / Interruption]
Quality Rating: [Error]
Brief Description: [The system stopped talking after I gave the feedback "OK."]
\end{lstlisting}
\caption{An example of a logged annotation case from the RW-HM dataset collection.}
\label{fig:annotation_case}
\end{figure*}

\subsection{Real-World Human-Human (RW-HH) Dataset}

\begin{itemize}
    \item \textbf{Data Source:} We use the publicly available portion of the \textbf{Seamless Interaction} dataset from Meta.
    \item \textbf{Filtering Pipeline for Interactive Segments:}
    \begin{enumerate}
        \item \textbf{Overlap Detection:} We employ the \texttt{pyannote.audio} toolkit for speaker diarization to identify all speech segments where both participants are speaking simultaneously.
        \item \textbf{Vocal Consistency Check:} For each detected overlap, we compare the timbre before and after the overlap using speaker embeddings to distinguish interruptions from backchannels.
        \item \textbf{Manual Verification:} A team of human annotators reviewed all automatically filtered segments to ensure quality and record interation events.
    \end{enumerate}
    \item \textbf{Annotation:} We compiled the meta information provided by the dataset and hired human experts to listen to the conversation segments and annotate the corresponding interaction events and fluency descriptions. Human-to-human interaction is a highly imbalanced dataset, and most conversations are fluent and natural, with few semantic errors.
\end{itemize}

\section{Model Training Details}
\label{train}
\subsection{Experimental Setup}
\begin{itemize}
    \item \textbf{GPU/CPU:} 4$\times$ NVIDIA H20 96GB GPUs
    \item \textbf{Operating System:} Ubuntu 22.04
    \item \textbf{Key Libraries:} PyTorch 2.6.0+cu124, Transformers 4.52.4, \texttt{ms-swift}, CUDA 12.4
\end{itemize}

\subsection{Training Hyperparameters}
Key hyperparameters for each stage are detailed in Table~\ref{tab:hyperparams}.

\begin{table*}[t]
\centering
\caption{Key hyperparameters for each stage of the training pipeline.}
\label{tab:hyperparams}
\begin{tabular}{@{}lccc@{}}
\toprule
\textbf{Parameter}              & \textbf{Stage 1 (SFT-1)} & \textbf{Stage 2 (SFT-2)} & \textbf{Stage 3 (GRPO)} \\
\midrule
Base Model                      & Qwen2.5-Omni            & SFT-1 Checkpoint         & SFT-2 Checkpoint        \\
Trained Components              & Full Model              & Full Model               & LLM Only (Encoder Frozen) \\
Learning Rate                   & 1e-5                    & 1e-5                     & 1e-6                    \\
Batch Size (per device)         & 4                       & 2                        & 2                       \\
Epochs                          & 2                       & 1                        & 1 ($\sim$400 steps)    \\
Optimizer                       & AdamW                   & AdamW                    & AdamW                   \\
Teacher Model                   & ---                     & Gemini-2.5-Pro           & ---                     \\
Candidate Samples (K)           & ---                     & ---                      & 4                       \\
Reward Weights ($\lambda$)      & ---                     & ---                      & $\lambda_{\text{fmt}}=0.5,\ \lambda_{\text{acc}}=0.5$ \\
Clip Range ($\epsilon$)         & ---                     & ---                      & 0.2                     \\
KL Penalty ($\beta$)            & ---                     & ---                      & 0.01                    \\
\bottomrule
\end{tabular}
\end{table*}

\subsection{Training Prompts and Instructions}
Our three-stage training pipeline utilizes distinct sets of instructional prompts tailored to the specific goal of each stage.

\subsubsection{Stage 1: Foundational Interaction Grounding (SFT-1)}
The goal of this stage is to teach the model the fundamental grammar of conversational structure. For each audio sample, we condition the model on one of three distinct tasks using a set of ten varied prompts per task.

\paragraph{Task 1: Interaction Event Identification}
The model is instructed to identify all predefined interaction events within the audio and output their type and timestamps.
\begin{itemize}
    \item \textbf{Example Prompts (10 variations used in training):}
    \begin{enumerate}
        \item Analyze the audio and identify all notable interaction events.
        \item List every interactional phenomenon, such as interruptions or backchannels.
        \item What interaction events occurred in this dialogue? Provide timestamps.
        \item Scan the conversation for turn-taking events and provide a log.
        \item Detect and timestamp all interruptions, backchannels, and pauses.
        \item Provide a list of all interactional events present in the recording.
        \item Identify key turn-management events from the audio.
        \item Report any instances of overlapping speech or significant silence.
        \item What is happening in this conversation from a turn-taking perspective?
        \item Log all communicative events beyond the speech content itself.
    \end{enumerate}
    \item \textbf{Example Ground-Truth Label:}
\begin{lstlisting}[language=json, basicstyle=\ttfamily\small, numbers=none]
[
  {"event_type": "Successful_Interruption", "start_time": "15.2", "end_time": "17.8"},
  {"event_type": "Backchannel", "start_time": "25.1", "end_time": "25.6"}
]
\end{lstlisting}
\end{itemize}

\paragraph{Task 2: Speaker Turn Segmentation}
The model must perform speaker diarization.
\begin{itemize}
    \item \textbf{Example Prompts (10 variations used in training):}
    \begin{enumerate}
        \item Diarize the following conversation.
        \item Segment the audio by speaker turn, providing timestamps for each.
        \item Who is speaking and when?
        \item Provide a speaker diarization log for the provided audio.
        \item Identify the start and end times for each speaker's utterance.
        \item Create a turn-by-turn breakdown of the dialogue.
        \item Which speaker is active at which timestamp?
        \item Segment the speech into discrete turns for Speaker A and Speaker B.
        \item Provide a complete speaker segmentation.
        \item Analyze the audio and output the speaker turn timeline.
    \end{enumerate}
    \item \textbf{Example Ground-Truth Label:}
\begin{lstlisting}[language=json, basicstyle=\ttfamily\small, numbers=none]
[
  {"speaker": "A", "start_time": "0.5", "end_time": "5.1"},
  {"speaker": "B", "start_time": "5.4", "end_time": "10.2"},
  {"speaker": "A", "start_time": "10.3", "end_time": "15.2"}
]
\end{lstlisting}
\end{itemize}

\paragraph{Task 3: Full Timestamped Transcription}
The model is instructed to generate a complete transcript.
\begin{itemize}
    \item \textbf{Example Prompts (10 variations used in training):}
    \begin{enumerate}
        \item Provide a full, timestamped transcript of the conversation.
        \item Transcribe the dialogue, including speaker labels and timestamps.
        \item Generate a detailed script of the conversation with timing information.
        \item What was said in the dialogue? Provide a complete, timed transcript.
        \item Create a transcription with speaker and time annotations.
        \item Transcribe the audio from start to finish with all details.
        \item Output the full text of the conversation with speaker turns and times.
        \item Convert the spoken dialogue into a timestamped text format.
        \item What is the full transcript of this interaction?
        \item Provide a verbatim transcription annotated with speaker and time data.
    \end{enumerate}
    \item \textbf{Example Ground-Truth Label:}
\begin{lstlisting}[language=json, basicstyle=\ttfamily\small, breaklines=true, numbers=none]
[
  {"speaker": "A", "start_time": "0.5", "end_time": "5.1", "text": "Hello, I wanted to ask about the return policy for an item I bought online."},
  {"speaker": "B", "start_time": "5.4", "end_time": "10.2", "text": "Of course, I can help with that. Do you have the order number handy?"}
]
\end{lstlisting}
\end{itemize}

\subsubsection{Stage 2: Distilling Interactional Reasoning (SFT-2)}
In this stage, the model learns to perform high-level, structured reasoning about interaction quality, guided by the comprehensive prompt shown in Figure~\ref{fig:prompt_sft2}.

\begin{figure*}[t]
\begin{lstlisting}[style=prompt, numbers=none]
# Interactional Dialogue Evaluation

**IMPORTANT**: Evaluation must include `<response think>` and `<fluency think>`
analysis and `<overall score>` rating.
Listen to a two-person interactional dialogue speech (Dual-channel audio,
with each channel representing one speaker), labeled as speakers A and B.
Evaluate the quality of the interaction, focusing on:
**Response Relevance:** 
**logical consistency, topic coherence**
**Interactional Fluency:**
**Detect and evaluate extended vocal overlaps, e.g., cross-channel overlap.**
**Detect and evaluate long pauses, e.g., pauses more than 3s between
speaker turns.

****Note**: Small pauses and brief overlaps in audio are acceptable, while
prolonged pauses and overlapping audio are harmful. You should consider
Response Relevance and Interactional Fluency separately, and provide the
corresponding thinking process.

## Scoring Criteria
Assign a single holistic score based on the combined evaluation:
`0` (Poor): Significant issues in either **Response Relevance ** or
**Interactional Fluency. **
`1` (Excellent): Both **Response Relevance ** and **Interactional Fluency **
are consistently appropriate and natural.
## Evaluation Output Format:
Strictly follow this template:
<response think>
[Analysing Response Relevance and giving reasons for scoring...]
</response think>
<fluency think>
[Analysing Interactional Fluency and giving reasons for scoring.]
</fluency think>
<overall score>X</overall score>
\end{lstlisting}
\caption{The  prompt used in Stage 2 (SFT-2)
, Stage 3 and evaluation.}
\label{fig:prompt_sft2}
\end{figure*}

\subsubsection{Stage 3: Performance and Generalization Enhancement (GRPO)}
The final stage uses Group Relative Policy Optimization. The base instruction prompt used during this stage to generate candidate evaluations is identical to the one used in SFT-2 (see Figure~\ref{fig:prompt_sft2}).

\section{COT Quality Evaluation Human Expert Evaluation Protocol}
\label{sec:human_eval_protocol}

To rigorously evaluate the quality, trustworthiness, and human alignment of the generated Chain-of-Thought (CoT) reasoning, we conducted a blind, human-centric evaluation. Qualified human experts were tasked with assessing the CoT outputs from our model and the two baselines. The identities of the models were anonymized and presented as 'Candidate A', 'Candidate B', and 'Candidate C' to prevent bias.

For each of the 40 test samples, experts were provided with the dialogue snippet, a pre-authored ground-truth rationale, and the three anonymized CoTs. They were instructed to perform the following two tasks sequentially.

\subsection{Task 1: Consistency Scoring}
The primary goal of this task is to independently assess how well each model's reasoning aligns with the ground-truth analysis.

\paragraph{Instructions for Experts:}
For each of the three candidate CoTs, you will perform the following steps:
\begin{enumerate}
    \item Carefully read the provided \textbf{Ground-Truth Rationale And Conversation Transcriptions}. This is the gold standard for your assessment.
    \item Read the \textbf{Generated CoT} from the candidate model.
    \item Evaluate the alignment between the two based on the following criteria:
    \begin{itemize}
        \item \textbf{Correctness:} Does the model correctly identify the key interaction events (e.g., interruptions, latencies, semantic errors) mentioned in the ground-truth rationale?~\cite{EN20120117X}
        \item \textbf{Completeness:} Does the model capture all the critical success or failure points detailed in the ground-truth rationale?
        \item \textbf{Reasoning Logic:} Is the model's justification for its final score logical and similar to the expert's reasoning path?
    \end{itemize}
    \item Assign a single \textbf{Consistency Score} on a 5-point Likert scale, where:
    \begin{itemize}
        \item \textbf{1:} Very Poor Alignment (Completely misses the key points or contradicts the rationale).
        \item \textbf{2:} Poor Alignment (Catches some minor points but misses the main issue).
        \item \textbf{3:} Moderate Alignment (Identifies the main issue but with incomplete or flawed reasoning).
        \item \textbf{4:} Good Alignment (Accurately reflects the ground truth with only minor omissions or differences).
        \item \textbf{5:} Perfect Alignment (The reasoning is a perfect match with the ground-truth rationale).
    \end{itemize}
\end{enumerate}

\subsection{Task 2: Comparative Selection (Best-Choice Task)}
The goal of this task is to determine which model produces the qualitatively superior reasoning in a direct, head-to-head comparison.

\paragraph{Instructions for Experts:}
After scoring all three candidates individually, you will now compare them against each other:
\begin{enumerate}
    \item Review the \textbf{Ground-Truth Rationale} and all three \textbf{Candidate CoTs} (A, B, and C) side-by-side.
    \item Select the \textbf{single best CoT} that most effectively analyzes the dialogue.
    \item Your decision should be based on the following criteria, in order of importance:
    \begin{enumerate}
        \item \textbf{Alignment with Rationale:} This is the most critical factor. The winning CoT must identify the same core issues or successes as the human expert. A model that correctly diagnoses a specific failure is superior to one that misses it, even if their final scores are the same.
        \item \textbf{Diagnostic Accuracy:} The reasoning must be sound. A model should not arrive at the right conclusion for the wrong reasons.
        \item \textbf{Clarity and Specificity:} The analysis should be clear, specific, and well-supported by evidence from the dialogue. Vague or generic statements are less valuable.
    \end{enumerate}
    \item Record the label of your chosen candidate (e.g., 'Candidate B').
\end{enumerate}

\section{Teacher Model Prompt for CoT Distillation}
\label{sec:cot_prompt}
To generate structured Chain-of-Thought (CoT) analyses from dialogue metadata, a powerful teacher model (e.g., a Gemini-class model) was guided by the prompt shown in Figure~\ref{fig:cot_prompt_detail}. This prompt instructs the model to produce a decoupled analysis of semantic coherence and interactional fluency, followed by a final binary score, based on a detailed dialogue record.

\begin{figure*}[h!]
\begin{lstlisting}[language=bash, numbers=none]
# SYSTEM INSTRUCTION

You are an expert in conversational analysis. Your task is to evaluate the overall interaction quality of a spoken dialogue based on its dual-channel transcript, speaker timestamps, and annotated interaction events.

Your analysis must be explicitly decoupled into two dimensions:
1.  **Response Relevance**: Evaluate the semantic appropriateness, coherence, and relevance of the assistant's response to the user's intent.
2.  **Interactional Fluency**: Evaluate the turn-taking mechanics and timing. Assess whether the assistant's behavior adheres to natural conversational norms, identifying issues such as inappropriate barge-ins, excessive latency, or ignored interruptions.

Finally, provide a single binary **Overall Score** (0 for failure, 1 for success) based on your comprehensive analysis.

## INPUT FORMAT:

You will receive a JSON object containing:
- `transcript`: A list of utterance objects, each with a speaker, start time, end time, and text.
- `interaction_events`: A list describing specific events (e.g., "Successful_Interruption") with types and timestamps.
- `error_type` (optional): The specific failure mode if the sample is a designed negative case.

## OUTPUT FORMAT:

You must strictly adhere to the following template without any additional explanatory text:

<response_think>
[Analyze the dialogue for content relevance and logical coherence. State whether the assistant's response addresses the user's query and is factually appropriate. If an interruption occurred, analyze if the assistant correctly processed the new information.]
</response_think>
<fluency_think>
[Analyze the dialogue for timing and turn-management fluency. Based on events and timestamps, determine if failures like Inappropriate Barge-in, Delayed Turn Transition, or Ignored Interruption are present. Justify your assessment with specific evidence from the transcript.]
</fluency_think>
<overall_score>[0 or 1]</overall_score>
\end{lstlisting}
\caption{Prompt for generating CoT annotations from dialogue metadata.}
\label{fig:cot_prompt_detail}
\end{figure*}

\section{Standardized Inference Prompt for Evaluation}
\label{sec:inference_prompt}
To ensure a fair and consistent evaluation across all baseline models in a zero-shot setting, the standardized prompt detailed in Figure~\ref{fig:prompt_sft2} was used.

\section{Full Training Sample Example}
\label{sec:training_sample}
Figure~\ref{fig:training_sample_detail} provides a complete training data sample. It illustrates a logically sound "Inappropriate Barge-in" scenario, where the assistant's response is a plausible (but incorrect) reaction based *only* on the pre-pause utterance fragment, thus not requiring any prescience.

\begin{figure*}[h!]
\begin{lstlisting}[language=json, numbers=none]
{
  "dialogue_metadata": {
    "transcript": [
      {
        "speaker": "User",
        "start_time": "1.2",
        "end_time": "8.5",
        "text": "Okay, I need to book a flight to... [pause] San Francisco for the conference next month."
      },
      {
        "speaker": "Assistant",
        "start_time": "6.1",
        "end_time": "9.3",
        "text": "[barge_in] I can certainly help with that. Where would you like to go?"
      }
    ],
    "interaction_events": [
      {
        "event_type": "Inappropriate_Barge_in",
        "start_time": "6.1",
        "end_time": "8.5",
        "participants": ["User", "Assistant"]
      }
    ],
    "error_type": "Inappropriate_Barge_in"
  },
  "model_output": {
    "chain_of_thought": "<response_think>\nThe assistant's response is semantically plausible given the user's initial utterance fragment ('I need to book a flight to...'). However, it becomes contextually redundant and inappropriate once the user completes their turn by specifying the destination. The assistant's question, 'Where would you like to go?', demonstrates a failure to process the full user utterance.\n</response_think>\n<fluency_think>\nThe interaction exhibits a critical fluency failure. The user's pause after 'to...' is a natural intra-turn hesitation for thought formulation, not a turn-yielding cue. The assistant incorrectly interprets this pause and begins its response prematurely at 6.1s, talking over the user who is still speaking until 8.5s. This constitutes a classic 'Inappropriate Barge-in' and severely disrupts the conversational flow.\n</fluency_think>\n<overall_score>0</overall_score>",
    "binary_score": 0
  }
}
\end{lstlisting}
\caption{A training sample for an 'Inappropriate Barge-in' error, where the model's response is based only on the pre-pause utterance fragment.}
\label{fig:training_sample_detail}
\end{figure*}

\end{document}